\documentclass[10pt,twocolumn,letterpaper]{article}

\usepackage{iccv}
\usepackage{times}
\usepackage{epsfig}
\usepackage{graphicx}
\usepackage{amsmath}
\usepackage{amssymb}
\usepackage{bbold}
\usepackage{color}
\usepackage{overpic}
\usepackage{makecell}
\usepackage{float}
\usepackage{wrapfig}
\usepackage{tabularx}
\usepackage{siunitx}
\usepackage{multirow}
\usepackage{textcomp}
\usepackage{pict2e}

\usepackage[accsupp]{axessibility}  

\usepackage[pagebackref=true,breaklinks=true,letterpaper=true,colorlinks,bookmarks=false]{hyperref}

\iccvfinalcopy 

\def\iccvPaperID{10402} 

\ificcvfinal\fi

\newcommand*\samethanks[1][\value{footnote}]{\footnotemark[#1]}
\begin{document}

\title{Sketch2Mesh: Reconstructing and Editing 3D Shapes from Sketches}

\author{Benoit Guillard\thanks{Equal contribution},  \hspace{7pt}Edoardo Remelli\samethanks,  \hspace{7pt}Pierre Yvernay,   \hspace{7pt}Pascal Fua\\\\
CVLab, EPFL\\
{\tt\small name.surname@epfl.ch}
}


\newif\ifdraft
\draftfalse
\drafttrue

\newcommand{\skm}{{\it Sketch2Mesh}}
\newcommand{\skmr}{{\it Sketch2Mesh/Render}}
\newcommand{\skmc}{{\it Sketch2Mesh/Chamfer}} 

\definecolor{orange}{rgb}{1,0.5,0}
\definecolor{violet}{RGB}{70,0,170}
\definecolor{pink}{RGB}{252,107,252}
\definecolor{brown}{RGB}{139,69,19}
\definecolor{red_fig}{RGB}{189,9,9}

 \newcommand{\PF}[1]{{\color{red}{\bf PF: #1}}}
 \newcommand{\pf}[1]{{\color{red} #1}}
 \newcommand{\ER}[1]{{\color{violet}{\bf ER: #1}}}
 \newcommand{\er}[1]{{\color{violet} #1}}
 \newcommand{\BG}[1]{{\color{blue}{\bf BG: #1}}}
 \newcommand{\bg}[1]{{\color{blue}{#1}}}
 \newcommand{\PY}[1]{{\color{orange}{\bf PY: #1}}}
 \newcommand{\py}[1]{{\color{orange}{#1}}}
 
\newcommand{\comment}[1]{}
\newcommand{\parag}[1]{\vspace{-3mm}\paragraph{#1}}
\newcommand{\sparag}[1]{\vspace{-3mm}\subparagraph{#1}}

\newcommand{\bp}{\mathbf{p}}
\newcommand{\bq}{\mathbf{q}}
\newcommand{\bv}{\mathbf{v}}
\newcommand{\bx}{\mathbf{x}}
\newcommand{\bz}{\mathbf{z}}

\newcommand{\bC}{\mathbf{C}}
\newcommand{\bF}{\mathbf{F}}
\newcommand{\bP}{\mathbf{P}}
\newcommand{\bV}{\mathbf{V}}
\newcommand{\bT}{\mathbf{T}}
\newcommand{\bX}{\mathbf{X}}
\newcommand{\bM}{\mathbf{M}}
\newcommand{\bN}{\mathbf{N}}

\newcommand{\cE}{{\cal E}}
\newcommand{\cD}{{\cal D}}
\newcommand{\cM}{{\cal M}}
\newcommand{\cR}{{\cal R}}

\maketitle
\ificcvfinal\thispagestyle{empty}\fi

\begin{abstract}

Reconstructing 3D shape from 2D sketches has long been an open problem because the sketches only provide very sparse and ambiguous information. In this paper, we use an encoder/decoder architecture for the sketch to mesh translation. When integrated into a user interface that provides camera parameters for the sketches, this enables us to leverage its latent parametrization to represent and refine a 3D mesh so that its projections match the external contours outlined in the sketch. We will show that this approach is easy to deploy, robust to style changes, and effective. Furthermore, it can be used for shape refinement given only single pen strokes. 

We compare our approach to state-of-the-art methods on sketches---both hand-drawn and synthesized---and demonstrate that we outperform them.

\end{abstract}



\section{Introduction}


\begin{figure}
\begin{center}
\begin{overpic}[clip, trim=0cm 4.0cm 13cm 0 cm,width= .45\textwidth]{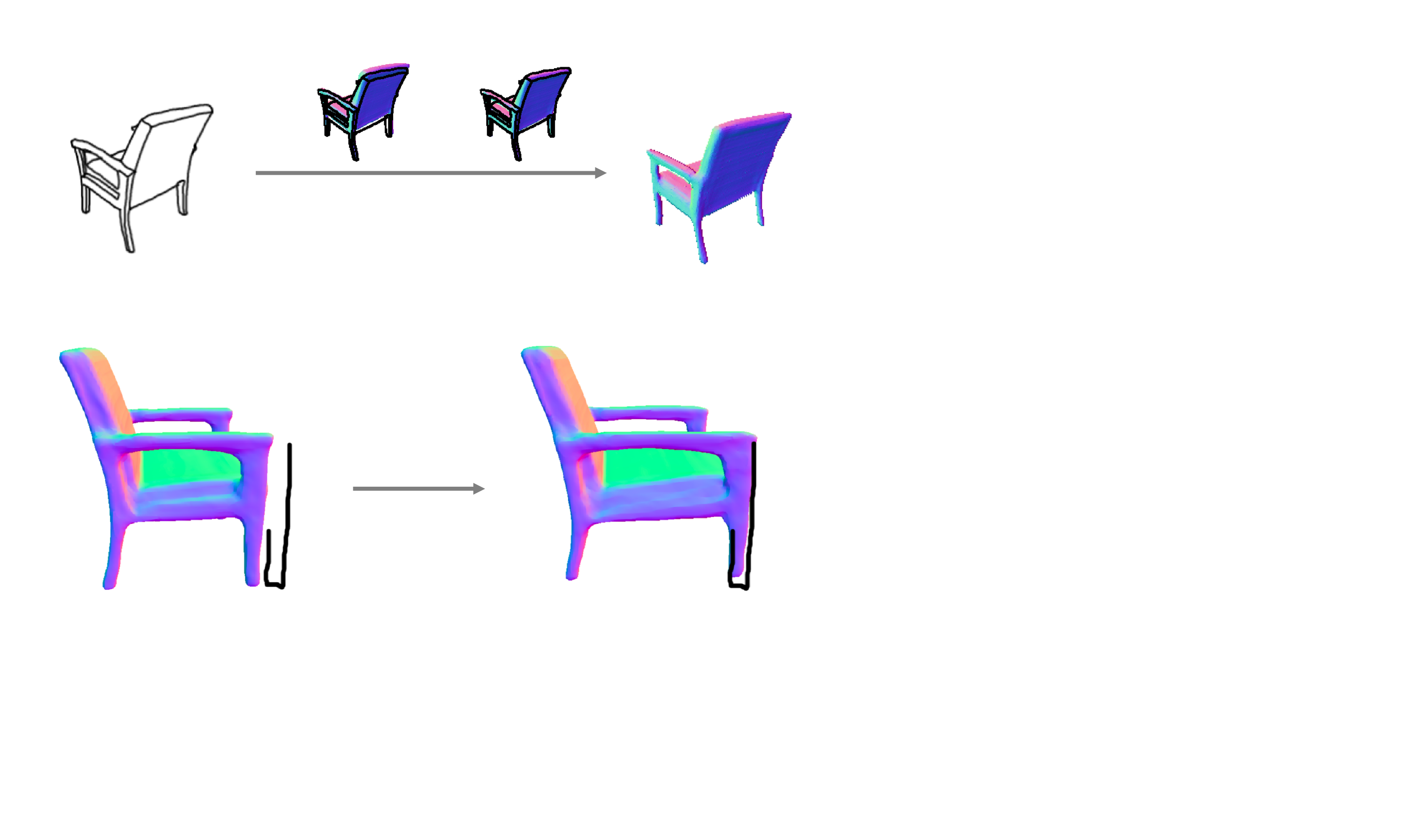}
\put(39,53){\small{(a) Reconstructing}}
\put(44,13){\small{(b) Editing}}
\end{overpic}
 \end{center}
%
%
%
%
%
%
\vspace{-2mm}
   \caption{\textbf{Sketch2Mesh.} We propose a pipeline for reconstructing and editing 3D shapes from line drawings. 
   	We train an encoder/decoder architecture to regress surface meshes from synthetic sketches.
    Our network learns a compact representation of 3D shapes that is suitable for downstream optimization: \textbf{(a)} When presented with sketches drawn in a style different from that of the training ones-- for example a real drawing -- aligning the projected external contours to the input sketch bridges the domain gap. \textbf{(b)} The same formulation can be used to enable unexperienced users to edit reconstructed shapes via simple 2D pen strokes. Best seen in Supplemental video.}
\label{fig:teaser}
\end{figure}

Reconstructing 3D shapes from hand-drawn sketches has the potential to revolutionize the way designers, industrial engineers, and artists interact with Computer Aided Design (CAD) systems. Not only would it address the industrial need to digitize vast amounts of legacy models, an insurmountable task, but it would allow practitioners to interact with shapes by drawing in 2D, which is natural to them, instead of having to sculpt 3D shapes produced by cumbersome 3D scanners.

Current deep learning approaches~\cite{Lun17a,Delanoy18,Song20b,Zhong20} that regress 3D point clouds and volumetric grids from 2D sketches have shown promise despite being trained on synthetic data, but yield coarse 3D surface representations that are cumbersome to edit. Furthermore, they require multi-view sketches for effective reconstruction~\cite{Delanoy18} or are restricted to a fixed set of views \cite{Lun17a}. 

Meanwhile Single View Reconstruction (SVR) approaches have progressed rapidly thanks to the introduction of new shape representations~\cite{Groueix18a,Park19c,Mescheder19,Remelli20b} along with novel architectures~\cite{Wang18e, Gkioxari19, Guillard20,Xu19b} that exploit image-plane feature pooling to align reconstructions to input images. Hence, it can seem like a natural idea to also use them for reconstruction from sketches. Unfortunately, as we will show, the sparse nature of sketch images makes it difficult for state-of-the-art SVR networks relying on local feature pooling from the image plane to perform well. This difficulty is compounded by the fact that different people sketch differently, which introduces a great deal of variability in the training process and makes generalization problematic.
Furthermore, these architectures do not learn a compact representation of 3D shapes, which makes the learned models unsuitable for down stream applications requiring a strong shape prior, such as shape editing.

To overcome these challenges, we train an encoder/decoder architecture~\cite{Remelli20b} to produce a 3D mesh estimate given an input line drawing. This yields a compact latent representation that acts as an information bottleneck. At inference time, given a previously unseen camera-calibrated sketch, we compute the corresponding latent vector and refine its components to make the projected 3D shape it parameterizes match the sketch as well as possible. In effect, this compensates for the style difference between the input sketch and those that were used for training purposes. We propose and investigate two different ways to do this:
\begin{enumerate}

 \item \skmr{}. We use a state-of-the-art image translation technique~\cite{Isola17} trained  to synthesize foreground/background images from sketches and then use the resulting images as targets for differentiable rasterization~\cite{Runz20,Remelli20b,Poursaeed20}.
 
 \item \skmc{}. We directly optimize the position of the 3D shape's projected contours to make them coincide with those of the input sketch by minimizing a 2D Chamfer distance. 
 
\end{enumerate}
Remarkably, \skmc{}, even though simpler, works as well or better than \skmr{}. The former exploits only {\it external} object contours for refinement purposes, which helps with generalization because most graphics designers draw these external contours in a similar way. It also makes it unnecessary the auxiliary network that turns sketches into foreground/background images.

A further strength of  \skmc{} is that it does not require backpropagation from a full rasterized image but only from sparse contours. Hence, it is naturally applicable for local refinement given a camera-calibrated partial sketch. And, unlike earlier work~\cite{Nealen05,Karpenko02,Kho05} on shape editing from local pen strokes it allows us to take into account a strong shape prior by relying on the latent vector, ensuring that shapes can be edited robustly with sparse 2D pen strokes.

\comment{However, this is hard to do because hand-drawn sketches do not obey rigid rules. They often include portions of the occluding contours that do not always define closed regions and the sparse nature of sketch images makes it difficult for even the best current single view reconstruction (SVR) networks, such as those of~\cite{Runz20,Remelli20b,Poursaeed20},  to yield accurate predictions. Furthermore, different people typically draw different contours inside the objects to denote details of interest. 

Current deep learning approaches~\cite{Lun17a,Delanoy18} that regress 3D volumetric grids from 2D sketches have shown promise despite being trained only on synthetic data, but yield coarse 3D surface representations that are cumbersome to edit. Furthermore, they require multi-view sketches for effective reconstruction~\cite{Delanoy18} or are restricted to a fixed set of views \cite{Lun17a}. \PF{Any newer ones?} \ER{Only one really, but it's gonna come out at ICLR21 so there is really no need to mention it?}\PF{Can you shoot it down?}

In this paper, we propose an approach that can refine an initial shape prediction given a single 2D sketch, is relatively insensitive to the drawing style, and does not require the training of an auxiliary network.  We represent the 3D object to be modeled by a 3D mesh parameterized in terms of a set of latent variables. These variables can be adjusted to deform the mesh so that its \bg{external} contours match the sketch as well as possible. In practice, we learn a latent shape representation for a class of objects and backpropagate through it to minimize the 2D chamfer distance of its \bg{external} contours to the sketch. 
}

\comment{


Recently, Single View Reconstruction (SVR) from RGB images has experienced tremendous progresses thanks to both the introduction of new shape representations \cite{groueix2018papier, guillard2020uclid, gkioxari2019mesh,Park19c,Mescheder19,xu2019disn}  and advances in differentiable rendering \cite{kato2018neural, liu2019soft, ravi2020accelerating}.
Can state-of-the-art SVR pipelines be readily used to reliably reconstruct 3D shapes from single view 2D line-drawings?

While seemingly similar to the task of SVR from RGB images, we observe that using sketches as input presents different challenges.
We show that the sparse nature of sketch images makes it difficult for state-of-the-art SVR networks to give accurate predictions. We suggest this is caused by their reliance on feature pooling from the image plane, which is mostly empty in the case of sketches.
To avoid this pitfall, we propose to use encoder/decoder architectures with a single vector code as low-dimensional bottleneck. 

Crucially, having a low-dimensional vector code representing a class of shapes opens the door to maximum a posteriori (MAP) optimization through differentiable rasterization (DR). \ER{In this work, we exploit this to refine network predictions at inference time, as well as to define a natural way for users to interact and edit reconstructed shapes locally through simple pen strokes.}

Specifically, recent work \cite{runz2020frodo,remelli2020meshsdf,Poursaeed20} has shown that 2D buffers -such as silhouettes or depth maps- can be used to refine 3D reconstructions produced by encoder/decoder architectures and thus allow networks trained on synthetic RGB renders to yield accurate reconstruction on real world images. These approaches rely on either estimating 2D buffers from input images using state-of-the-art segmentation/depth estimation networks trained on large-scale real world datasets \cite{Lin14a}, or acquiring the additional information through specific sensors.

As requiring users to manually provide a binary segmentation mask is cumbersome, applying refinement techniques to line-drawings  would require us to use an auxiliary network to infer occupancy masks from input sketches. Due to the lack of large-scale line-drawings datasets \cite{song2020deep}, however, we found that such networks struggle at generalizing to different sketching styles, and thus make refinement through differentiable rasterization less effective, or in some cases detrimental.

By contrast, we propose a refinement method that does not require training an auxiliary network. Acting directly on mesh surface points, it stirs them towards external contours present in the sketch, and does not rely on any estimate of 2D occupancy mask. We demonstrate the greater robustness to sketching style variation of our method compared to DR methods, and its applicability to various mesh based representations. \PF{How is the mesh initialized? From the sketch?}

Importantly, since our method does not rely on backpropagation from a full rasterized image, it is naturally applicable for local refinement using a partial sketch. Specifically, we demonstrate it is suitable for interactive shape editing, via a sketching interface we developed.

}
\comment{
We study the reconstruction of 3D shapes from sketches. While seemingly similar to the task of single view reconstruction (SVR) from RGB images, using sketches as input images is challenging in different ways. We show that the sparse nature of sketch images makes it difficult for state of the art SVR networks to give accurate predictions. We suggest this is caused by their reliance on feature pooling from the image plane, which are mostly empty in the case of sketches.

To avoid this pitfall, we propose to use an encoder/decoder architecture with a single vector code as bottleneck. Moreover, having a single vector code representing a shape opens to door to maximum a posteriori optimization based on differentiable rendering techniques. In this context, classical optimization via differentiable rasterization require using auxilliary networks to infer occupancy masks or normal maps from sketches. This makes differentiable rendering refinement less effective in the presence of a data domain gap - for example a change in sketching style.

By contrast, we propose a refinement method that does not require training an auxilliary network. Acting directly on surface points, it stirs them towards external contours present in the sketch, and does not rely on any estimate of surface normals nor 2D occupancy mask. We demonstrate the greater robustness to sketching style variation of our method compared to DR methods.

Finally, since our method does not rely on backpropagation from a full rasterized image, it is applicable for local refinement using a partial sketch. We demonstrate it is suitable for interactive shape editing, via a sketching interface we developed.
}


\section{Related Work}

Recent years have seen an explosion in 3D shape modeling capabilities from images in general and sketches in particular. In this section, we first review some of the new shape representation methods that have made this possible and then discuss how specific advances relate to the method we propose. 

\parag{Surface Representation.}

Among existing 3D surface representations, meshes made of vertices and faces are one of the most popular and versatile types and many early surface-modeling methods focused on deforming pre-existing templates based on such meshes that were either limited by design to a fixed topology~\cite{Fua96f,Salzmann09a} or required {\it ad hoc} heuristics that do not generalize well~\cite{McInerney99}. Furthermore, because meshes can have variable numbers of vertices and facets, it is challenging to make this representation suitable to deep learning architectures. A standard approach has therefore been to use graph convolutions to deform a pre-defined template \cite{Monti17, Wang18e}. Hence, it is limited to a fixed topology by design. A promising alternative~\cite{Groueix18a} is to use a union of surface patches instead, which can handle arbitrary topologies. However, this method does not offer any guarantee that patches stitch together correctly and, in practice, yields non watertight surfaces. 


Another alternative is to use an implicit description where the surface is described by the zero crossings of a volumetric function $\Psi: R^3 \rightarrow R$ ~\cite{Sethian99} whose values can be adjusted. The strength of this implicit representation is that the zero-crossing surface can change topology without explicit re-parameterization. Until recently, its main drawback was thought to be that working with volumes, instead of surfaces, massively increased the computational burden. 


This changed dramatically two years ago with the introduction of continuous deep implicit-fields. They represent 3D shapes as level sets of deep networks that map 3D coordinates to a signed distance function~\cite{Park19c} or an occupancy field~\cite{Mescheder19,Chen19c}. This mapping yields a continuous shape representation that is lightweight but not limited in resolution. 

However, for applications requiring explicit surface parameterizations, the non-differentiability of standard approaches to iso-surface extraction~\cite{Lorensen87} remains an obstacle to exploiting the advantages of implicit representations. This was overcome recently by introducing a differentiable way to produce explicit surface mesh representations from Deep Signed Distance-Functions~\cite{Remelli20b}. It was shown that, by reasoning about how implicit-field perturbations affect local surface geometry, one can differentiate the 3D location of surface samples with respect to the underlying deep implicit-field. This insight resulted in the {\it  MeshSDF} end-to-end differentiable architecture that takes as input a compact latent vector and outputs a 3D watertight mesh and that we use here.


\begin{figure*}[t]
\begin{center}
\begin{tabular}{ccc}
 \includegraphics[width=.20\textwidth]{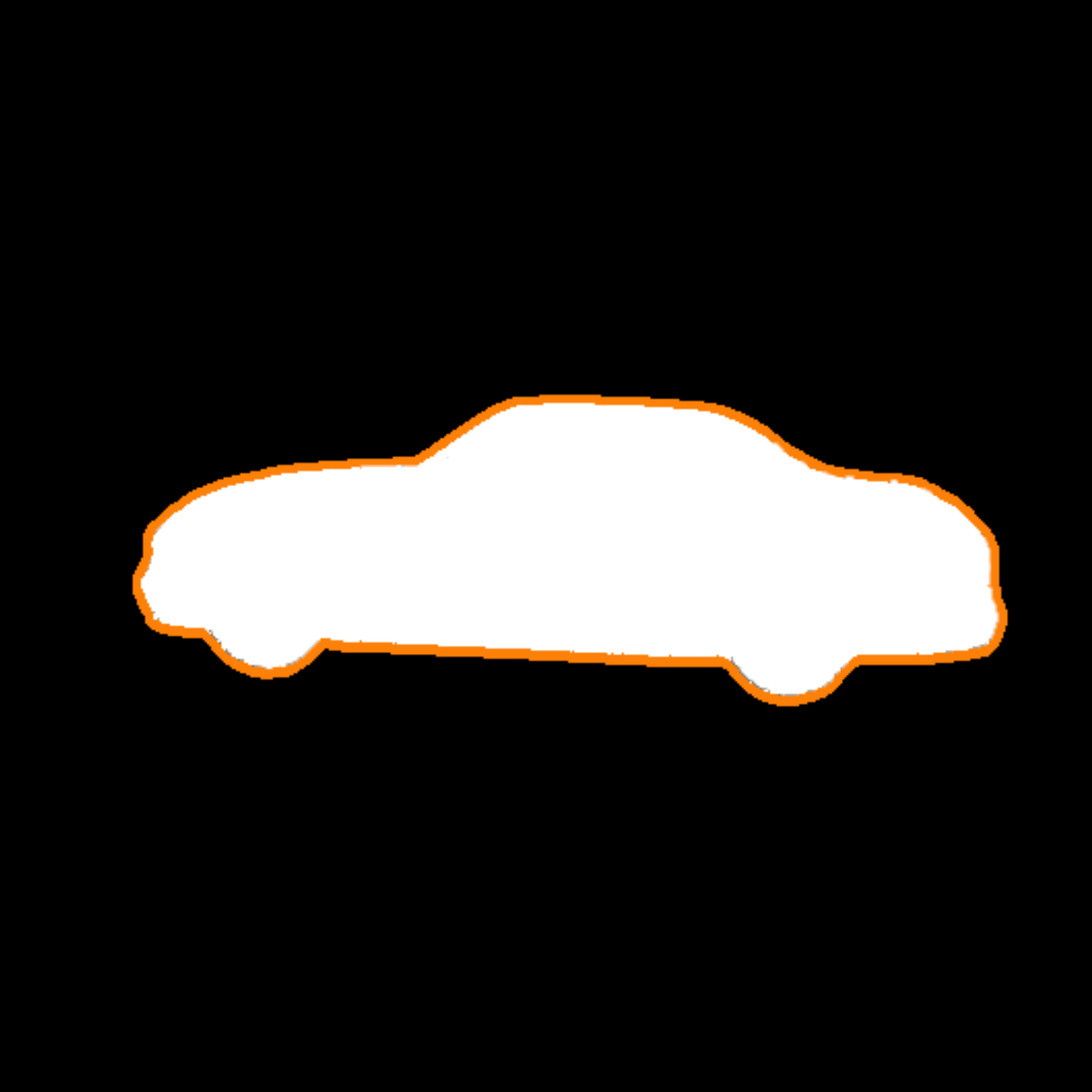}&
 \includegraphics[width=.35\textwidth]{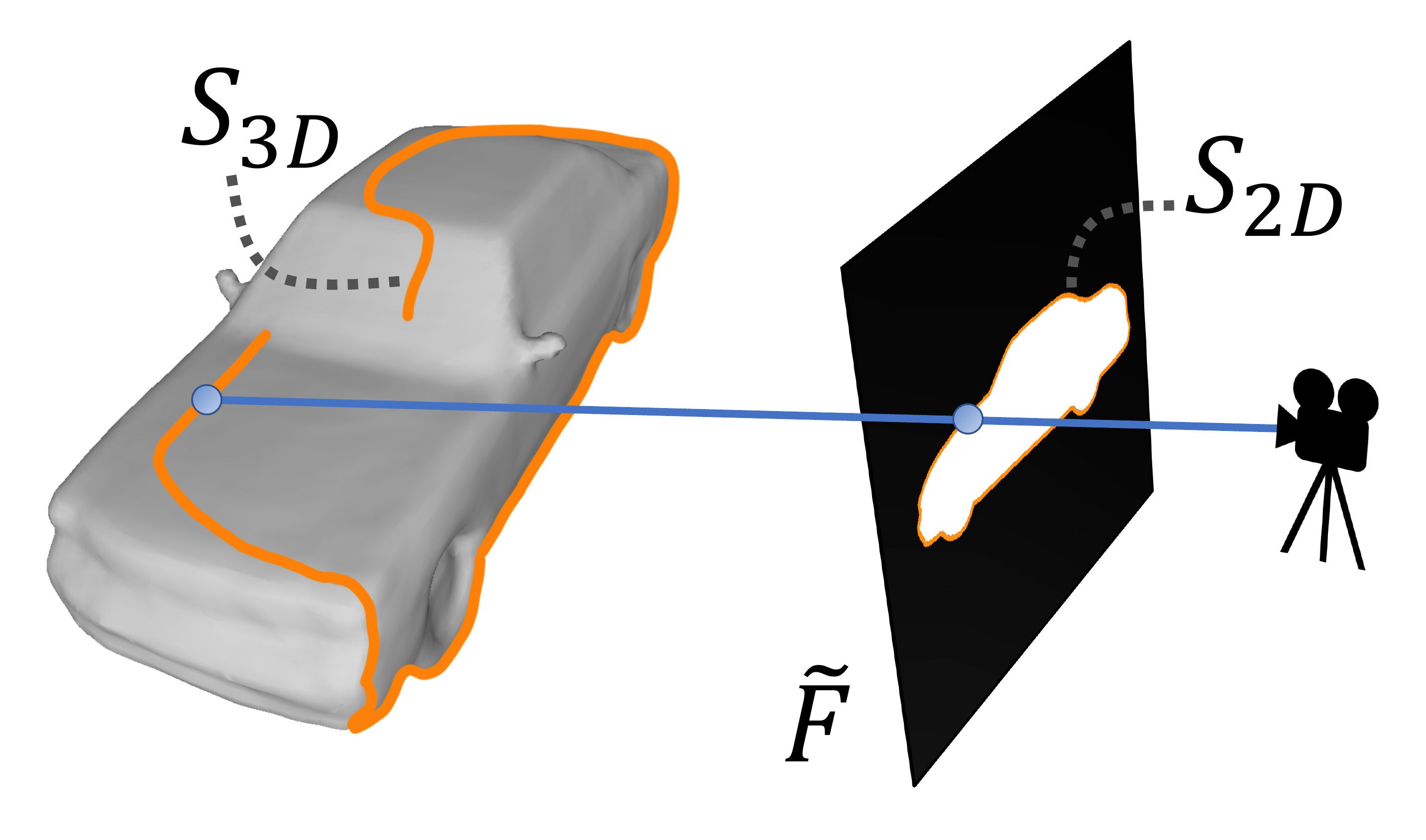} &
 \includegraphics[width=.20\textwidth]{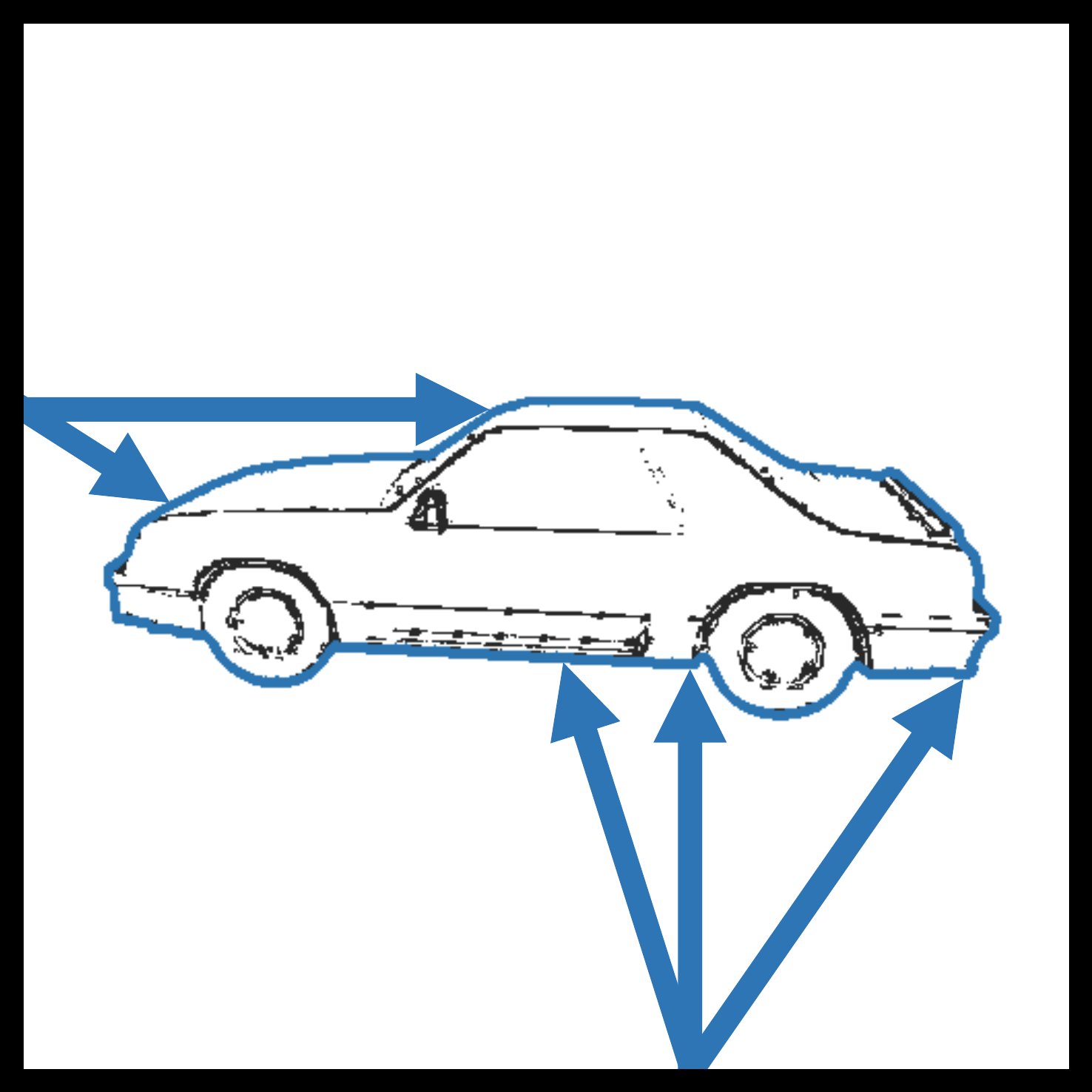} \\
    (a) & (b) & (c)
\end{tabular}
\end{center}
\vspace{-3mm}
   \caption{\textbf{External contours in 2D and 3D.} \textbf{(a)} The external contours of the projected mesh are shown in orange. They form the $S_{2D}$ set of  Eq.~\ref{eq:s2d}. \textbf{(b)} The corresponding 3D points on the mesh are also overlaid in orange. They form the $S_{3D}$ set of  Eq.~\ref{eq:s3d}. \textbf{(c)} We filter the original sketch to keep only the external contours, which will be matched against $S_{2D}$.}
\label{fig:method}
\end{figure*}
\parag{3D Reconstruction from Sketches.} 

Reconstructing 3D models from line drawings has also been an active research area for more many decades. Early attempts tackled the inherent ambiguity of this inverse problem by either assuming that the drawn lines represent specific shape features \cite{Malik89,Igarashi06} or by constraining the class of 3D shapes that can be handled~\cite{Leclerc92,Lipson96,cordier2013inferring,jung2015sketching}.  More recently inflatable surface models~\cite{Dvorovzvnak20} demonstrated easy animation of the reconstructed shapes, but still constrain the artist to draw from a side view of the object and are limited to a fixed topology.
The emergence of deep learning has given rise to models \cite{Lun17a,Delanoy18,Jin20b} that can be far more expressive and have therefore boosted both the performance and generalization of algorithms that parse sketches into 3D shapes. Given an input sketch, \cite{Lun17a} regress depth and normal maps from 12 viewpoints, and fuse them to obtain a dense point cloud from which a surface mesh is extracted. Their pipeline, however, must be trained for each input sketch viewpoint, making it incompatible with a free viewpoint sketching interface. In~\cite{Delanoy18}, a 3D convolutional network trained on a catalog of simple shape primitives regresses occupancy grids from sketches. In addition to the limited output resolution, a refinement strategy based on sketches from multiple views is needed for effective reconstruction. \cite{Jin20b} jointly projects 3D shapes and their front, side and top views occluding contours in the embedding space of a VAE. Their pipeline is trained on a single sketch style (occluding contours) and outputs volumetric grids. At inference time it retrieves the closest embedding code that was seen during training, thus limiting its generalization capabilities. 

\parag{Single View Reconstruction.}

Recently, Single View Reconstruction (SVR) from RGB images has also experienced tremendous progresses thanks to both the introduction of new shape representations discussed above and to he introduction of new SVR architectures \cite{Wang18e, Gkioxari19, Guillard20,Xu19b} relying on image-plane feature pooling to align reconstructions to input images. Unfortunately, many of these methods rely on feature pooling and therefore lack a compact latent representation that can be used for downstream applications that require strong shape priors, such as refinement or editing.  However, there are SVR methods that feature compact surface representations and we discuss below those that leverage either differentiable rendering or contours, as we do.


\sparag{Refinement using differentiable rendering.}

Recent work \cite{Runz20,Remelli20b,Poursaeed20} has shown that 2D buffers -such as silhouettes or depth maps- can be used to refine 3D reconstructions produced by encoder/decoder architectures and thus allow networks trained on synthetic RGB renders to yield accurate reconstruction on real world images. These approaches rely on either estimating 2D buffers from input images --- using state-of-the-art segmentation/depth estimation networks trained on large-scale real world datasets \cite{Lin14a} --- or acquiring the additional information through specific sensors. Applying refinement techniques to line-drawings would require to use an auxiliary network to infer occupancy masks from input sketches. However we found that such networks struggle at generalizing to different sketching styles. This is due to the lack of diversified large-scale line-drawings datasets \cite{Gryaditskaya19}, and makes refinement through differentiable rasterization less effective, or in some cases detrimental. 

\sparag{Refinement by matching Silhouettes.}

Silhouettes have long been used to track articulated and rigid objects by modeling them using volumetric primitives whose occluding contours can be computed given a pose estimate. The quality of these contours can then be evaluated using either the chamfer distance to image edges~\cite{Gavrila95} or more sophisticated measures~\cite{Sminchisescu03a, Agarwal04a}.  Other approaches to exploiting external contours rely on minimizing the distance between the 3D model and the lines of sights defined by these contours~\cite{Ilic07a}. Our approach follows this tradition but combines silhouette alignment to a far more powerful latent representation.



\comment{
	
Reconstructing 3D models from line drawings has been an active research area for more than three decades. Early attempts tackled the inherent ambiguity of this inverse problem by either assuming that the lines in drawings represent specific shape features \cite{Malik89,Igarashi06} or by constraining the class of 3D shapes that can be handled~\cite{Leclerc92,Lipson96,cordier2013inferring,jung2015sketching}. 
Silhouettes have also long been used to track  articulated and rigid objects by modeling them using volumetric primitives whose occluding contours can be computed given a pose estimate. The quality of these contours can then be evaluated using either the chamfer distance to image edges~\cite{Gavrila95} or more sophisticated measures~\cite{Sminchisescu03a, Agarwal04a}.  Other approaches to exploiting external contours rely on minimizing the distance between the 3D model and the lines of sights defined by these contours~\cite{Ilic07a}. \bg{Recent inflatable surface models~\cite{Dvorovzvnak20} allow for easy animation of the reconstruted shapes, but constrain the artist to draw from a side view of the object.}

\subsection{Deep Modeling from Sketches} 

For both reconstruction and tracking as described above, strong models were required to make sense of the sparse 3D information provided by 2D sketches. The emergence of deep learning has given rise to newer models that can be far more expressive and have therefore boosted the performance of algorithms that parse sketches into 3D shapes. 

\PF{Describe here~\cite{Lun17a,Delanoy18,Jin20b,Dvorovzvnak20} and their limitations that you will remove.}

\bg{Given an input sketch, \cite{Lun17a} regress depth and normal maps from 12 viewpoints, and fuse them to obtain a dense point cloud. Their pipeline must be trained for each input sketch viewpoint, making it incompatible with a free viewpoint sketching interface. In~\cite{Delanoy18}, a 3D convolutional network trained on a catalog of simple shape primitives regresses volumetric grids from sketches. In addition to the limited output resolution, a refinement strategy based on sketches from multiple views is needed for effective reconstruction. \cite{Jin20} jointly project 3D shapes and their front, side and top views occluding contours in the embedding space of a VAE. Their pipeline is trained on a single sketch style (occluding contours) and outputs volumetric grids. At inference time it retrieves the closest embedding code that was seen during training, thus limiting its generalization capabilities.}

\PF{Section could end here.}

\ER{copy and paste from recent paper:

Lun et al. [23] predict
normal and depth maps as seen from 12 viewpoints, that
are fused to a dense point cloud. Li et al. [17] target free
form surfaces and introduced the intermediate layer that
predicts dense curvature directions. The method supports
sparse labels for depth maps and curvature hints for strokes.
It makes assumptions on line rendering: e.g., the silhouette
lines are assumed to be sketched in black and other lines in
grey

In~\cite{Delanoy18} a U-Net~\cite{Ronneberger15} based architecture, where they
encode 3D shape in a voxel representation and estimate the
probability of each voxel to be occupied. Their method
exploits the dedicated sketching interface, and their multiview sketching shape update strategy relies on the known
perspective camera-parameters. 

Jin et al. [15] learn the embedding of the shape given silhouettes of the 3D shape from
the front, side and top views. While it only exploits the information contained in the shape silhouettes, it proposes an
interesting idea for single sketch modeling of retrieving the
two additional views in the embedded space, prior to 3D
reconstruction.}

\textit{3D Shape Reconstruction from Sketches via Multi-view Convolutional Networks (2018 3DV)}: Multiview fusion to reconstruct shape, but need one shape encoder per input view (1 for top views, 1 for front views...)

\textit{Delannoy}: volumetric grid (=voxels), trained on catalog of simple shape primitives

\textit{Sorkine - Monster Mash}: use inflatable surfaces, and no category specific prior to create animatable 3D shapes from sketches. But for inflation to work, the provided sketch needs to be in profile view.

\textit{A Sketch-Based Interface for Detail-Preserving Mesh Editing (2005 Siggraph)}: allows local refinement of 3D shapes from partial sketches, but requires the user the define explicit handles, use no shape prior (moving one car wheel will not mode the other), need to explicitly regularise the mesh to ensure it remains nicely triangulated (we do not need to do this).

\BG{We also use/benchmark agains the following SVR methods. Should we present them here, or only in the experiments section?}

\paragraph{Single view reconstruction from RGB.}
\textit{DISN}: reconstruct SDFs, using view aligned local features pooled from 2D feature maps.

\textit{Mesh-RCNN}, reconstruct mesh with free topology.

\textit{UCLID-Net}, reconstruct surface patches. Patches are disconnected but free topology.

None of the above allows refinement through the learned generative shape model, since there is no compact code to optimize, as opposed to:

\textit{AtlasNet}: learn code to surface patches

\paragraph{Shape optimization through differentiable rendering.}
Maybe having a few references on this topic would be good too?

}


\section{Method}
\label{sec:method}
\subsection{Formalization}

Let $\bC \in \{0,1\}^{H\times W}$ be a binary image representing a sketch and let $\Lambda : \mathbb{R}^3 \rightarrow \mathbb{R}^2$ denote the function that projects 3D points into that image. By convention,  $\bC[i,j]$ is $0$ if it is marked by a pen stroke, and $\bC[i,j] = 1$ otherwise. 

We learn an encoder $\cE$ and a decoder $\cD$ such that $\cD \circ \cE(\bC) $ yields a mesh $\cM_{\Theta} = (\bV_{\Theta}, \bF_{\Theta})$.  $\Theta = \cE(\bC)$ is the latent vector that parameterizes our shapes. $\bV_{\Theta}$ and $\bF_{\Theta}$ represent the 3D vertices and facets. In practice, we use the MeshSDF encoding/decoding network architecture of~\cite{Remelli20b}. In general, $\cM_{\Theta}$ represents a 3D shape whose projection $\Lambda(\cM_{\Theta})$ only roughly matches the sketch $\bC$. Hence, our subsequent goal is to refine $\Theta$ so as to improve the match. 

We can achieve this in of two ways. We can turn the sketch into a foreground/background image and use differentiable rasterization to ensure that the projection of $\cM_{\Theta}$ matches that image. Alternatively, we can minimize the 2D Chamfer distance between the sketch and the projection. We describe both alternatives below. 

\subsection{Using Differential Rendering}
\label{sec:rendering}

In this method that we dubbed  \skmr{}, we train an image translation technique~\cite{Isola17} to synthesize foreground/background images from sketches. We denote as  $\bM \in \{0,1\}^{H\times W}$ this foreground/background image estimated from the input sketch $\bC$. On the other hand, we use the differentiable rasterizer~\cite{ravi2020accelerating}  $\cR^{F/B}$ to render a foreground/background mask $\widetilde{\bM} = \cR^{F/B}_{\Lambda}(\cM_{\Theta})$ of the projection of $\cM_{\Theta}$ by $\Lambda$. In $\widetilde{\bM}$, a pixel value is 1 if it projects to the surface of the mesh $\cM_{\Theta}$, and 0 otherwise. Finally, we refine $\cM_{\Theta}$  shape by minimizing
\begin{small}
  \begin{equation}
  \mathcal{L}_{F/B} = \left \|  \bM - \widetilde{\bM} \right \|^2 \; ,
  \end{equation}
\end{small}
the $L_2$ difference between $\bM$ and $\widetilde{\bM}$ with respect to $\Theta$. 

While conceptually straightforward, this approach is in fact quite complex because it depends on two off-the-shelf  but complex pieces of software, the rasterizer~\cite{ravi2020accelerating} and image-translator~\cite{Isola17}, one of which has to be trained properly.  
We now turn to a simpler technique that can be implemented from scratch and does not rely on an auxiliary neural network.

\subsection{Minimizing the 2D Chamfer Distance}
\label{sec:chamfer}

The simpler  \skmc{} approach involves directly finding those 3D mesh points that project to the contour of the foreground image and then minimizing the Chamfer distance between this contour and the sketch.

\subsubsection{Finding External Contours in 2D and 3D}
\label{sec:contours}

To identify surface points on $\cM_{\Theta}$ that project to exterior contour pixels, we first use $\Lambda$ to project the whole mesh onto a ${H\times W}$ binary image $\widetilde{\bF}$ in which all pixels are one except those belonging to external contours, such as those shown in orange in Fig.~\ref{fig:method}(a). Then, for each zero-valued pixel $\bp$ in $\widetilde{\bF}$, we look for a 3D point $\bP$ on one of the mesh facets that projects to it, that is, a point that is visible and such that $\Lambda(\bP)=\bp$. In theory, this can be done by finding to which facet $\bp$ belongs and then computing the intersection between the line of sight and the plane defined by that facet. In practice, we use Pytorch3d~\cite{ravi2020accelerating} which provides us with the facet number along with the barycentric coordinates of $\bP$ within that facet. Hence, we write
\begin{equation} \label{eq:bary_coords}
\bP = \alpha_1 \bV_1 + \alpha_2 \bV_2 + \alpha_3 \bV_3 \; ,
\end{equation}
with $\bV_1$, $\bV_1$ and $\bV_3$ are the vertices of the fact to which  $\bP$ belongs and $\alpha_1 + \alpha_2 + \alpha_3 = 1$. Since the coordinates of the three vertices are differentiable functions of $\Theta$, so are those of $\bP$. Repeating this operation for all external contour points yields a set of 3D points $S_{3D}$ such that
\begin{equation}
  \forall \bP \in S_{3D} \quad \widetilde{\bF}[\Lambda(\bP)] = 0 \; ,  \label{eq:s3d}
  \end{equation}
 along with a corresponding set of 2D projections
\begin{equation}
 S_{2D} = \{ \Lambda(\bP) | \bP \in S_{3D}\} \; . \label{eq:s2d}
  \end{equation}
Fig.~\ref{fig:method}(b) depicts such a set.

\subsubsection{Objective function}
\label{sec:obj}

To exploit the target sketch $\bC$, we first filter it to only preserve external contours. To this end, we shoot rays from the 4 image borders and only retain the first black pixels hit by a ray, as shown in Fig. \ref{fig:method}(c). This yields a filtered sketch $\bF \in \{0,1\}^{H \times W}$. As before,  $\bF[\bp]=0$ for pixels $\bp$ belonging to external contour and $\bF[\bp]=1$ for others. The ray-shooting algorithm we use is described in details in the supplementary material.

Our goal being for $\bF$, the filtered sketch, and $\widetilde{\bF}$, the external contours of the projected triangulation introduced in Section~\ref{sec:contours}, to match as well as possible, we write the objective function to be minimized as the bidirectional 2D Chamfer loss
\begin{small}
\begin{equation}
\mathcal{L}_{CD} =
\! \! \! \! \sum_{\mathbf{u} \in S_{2D}} \min_{\mathbf{v} | \bF[\mathbf{v}]=0} \! \left \|  \mathbf{u} - \mathbf{v} \right \|^2
+
\! \! \! \! \sum_{\mathbf{v} | \bF[\mathbf{v}]=0} \min_{\mathbf{u} \in S_{2D}} \! \left \|  \mathbf{u} - \mathbf{v} \right \|^2 \; . \label{eq:chamfer}
\end{equation}
\end{small}
The coordinates of the 3D vertices in $S_{3D}$ are differentiable with respect to $\Theta$. Since $\Lambda$ is differentiable, so are their 2D projections in  $S_{2D}$ and $\mathcal{L}_{CD}$ as whole.

\subsection{Using a Partial Sketch}
\label{sec:partial}

Minimizing the 2D Chamfer distance between external contours as described above does not require the input sketch to depict the shape in its entirety. This enables us to take advantage of partial sketches made of a single stroke. In this case, we can simply take the filtered sketch $\bF$ introduced above to be the sketch itself. But we must ensure that parts of the surface which project far away from the sketch remain unchanged. The rationale for this is that the initial shape should be preserved except where modifications are specified. To this end, we regularize the refinement procedure as follows. 

Given the initial value $\Theta_0$ of the latent vector we want to refine along with differentiable rasterizers~\cite{ravi2020accelerating}  $\cR^N$ and $\cR^{F/B}$ that return the normal maps $\bN_{\Theta}$ and foreground/background mask $\bM_{\Theta}$ given mesh $\cM_{\Theta}$, respectively, we minimize 
\begin{small}
\begin{align}
\mathcal{L}_{partial} & = \mathcal{L}_{CD} \\
  & + 
  \left \| \mathbb{1}_{t}  \circ (\bM_{\Theta} - \bM_{\Theta_0}) \right \|^2 +
  \left \| \mathbb{1}_{t} \circ (\bN_{\Theta} - \bN_{\Theta_0})  \right \|^2 \nonumber
\end{align}
\end{small}
where $\mathcal{L}_{CD}$ is the Chamfer distance of Eq.~\ref{eq:chamfer},  $\mathbb{1}_{t}$ is a mask that is zero within a distance $t$ of the sketch and one further away, and $\circ$ is the element wise product. In other words, the parts of the surface that project near to the sketch should match it and the others should keep their original normals and boundaries. 

Crucially, this is something that could not be done using the approach of Section~\ref{sec:rendering}, which requires complete sketches. This comes at the cost of having to use a differential renderer, unlike the approach of Section~\ref{sec:chamfer}. But this still does not require a trained network for image translation, which makes it easy to deploy.


\comment{
Moreover, 3D surface points $S_{3D}$ are projected to 2D image coordinates, using known camera $\Lambda$. It thus creates a 2D point cloud $S_{2D}$, which is the projection of $S_{3D}$ to the image plane:
\begin{equation}
S_{2D} = 
\end{equation}
Since projection $\Lambda$ is a differentiable operation, $S_{2D}$ is itself differentiably parametrized by $\Theta$. It contains the 2D points located on the exterior contours of $\cM_{\Theta}$ when visualized from $\Lambda$, ie. the coordinates of $0$-valued pixels of $\widetilde{\bF}$. While $\widetilde{\bF}$ is a raster image that was not rendered differentiably, $S_{2D}$ can be expressed directly as a function of $\bV_{\Theta}$. Our refinement objective stirs this point cloud towards the exterior contours of the target filtered sketch $\bF$, using a bidirectional 2D Chamfer loss:
}

\comment{

First, we render a synthetic sketch of $\cM_{\Theta}$ from the viewpoint of $\Lambda$, using the sketch renderer presented in (experiments section), and get a raster image $\widetilde{\bC}  \in \{0,1\}^{H\times W}$. This image is transformed into $\widetilde{\bF}$ by only keeping the most exterior sketch strokes. To do so, we shoot rays from the 4 image borders and only preserve the first black pixels hit by a ray, as shown in Fig. \ref{fig:method}(a). Strokes in $\widetilde{\bF}$ are therefore a subset of the ones in $\widetilde{\bC}$, and are meant to represent an approximation of the silhouette and discard sketch styles variations, which mostly arise among inner strokes.

Then, we back-project contour pixels in $\widetilde{\bF}$ to obtain a set of 3D points $S_{3D}$ on the mesh $\cM_{\Theta}$ that project exactly to these pixels, as shown in Fig. \ref{fig:method}(b). This translates into two conditions: (1) each 3D point $\mathbf{p} \in S_{3D}$ is visible and projects to an exterior contour pixel in $\widetilde{\bF}$, ie. $\widetilde{\bF}[\Lambda(\mathbf{p})] = 0$, and conversely (2) for each contour pixel $(i,j)$ of $\widetilde{\bF}$ (ie. $\widetilde{\bF}[i,j] = 0$) there is a point $\mathbf{p} \in S_{3D}$ such that $\Lambda(p) = (i,j)$.

Note that the above steps for selecting a subset of points on the surface (rendering of a sketch, filtering of external contours, and back-projection) are \textit{not differentiable} with regards to $\Theta$. To allow backpropagation to $\Theta$ in later stages of our method, the 3D points in $S_{3D}$ are expressed in barycentric coordinates of the triangle face they belong to. For each 3D point $\mathbf{p} \in S_{3D}$, there exists 3 scalars $(\alpha_1, \alpha_2, \alpha_3)$ and 3 vertices $(\mathbf{v}_1, \mathbf{v}_2, \mathbf{v}_3) \in \bV_{\Theta}$ such that
\begin{equation}
\mathbf{p} = \alpha_1 \mathbf{v}_1 + \alpha_2 \mathbf{v}_2 + \alpha_3 \mathbf{v}_3
\end{equation}

Since vertices positions are a differentiable function of $\Theta$, expressing points as linear combination of vertices as in Eq. \ref{eq:bary_coords} makes $S_{3D}$ differentiably parametrized by $\Theta$ as well.
}


\section{Results}


\begin{figure}
\begin{center}
	\begin{overpic}[clip, trim=0cm 3.0cm 3.5cm 0cm,width= .45\textwidth]{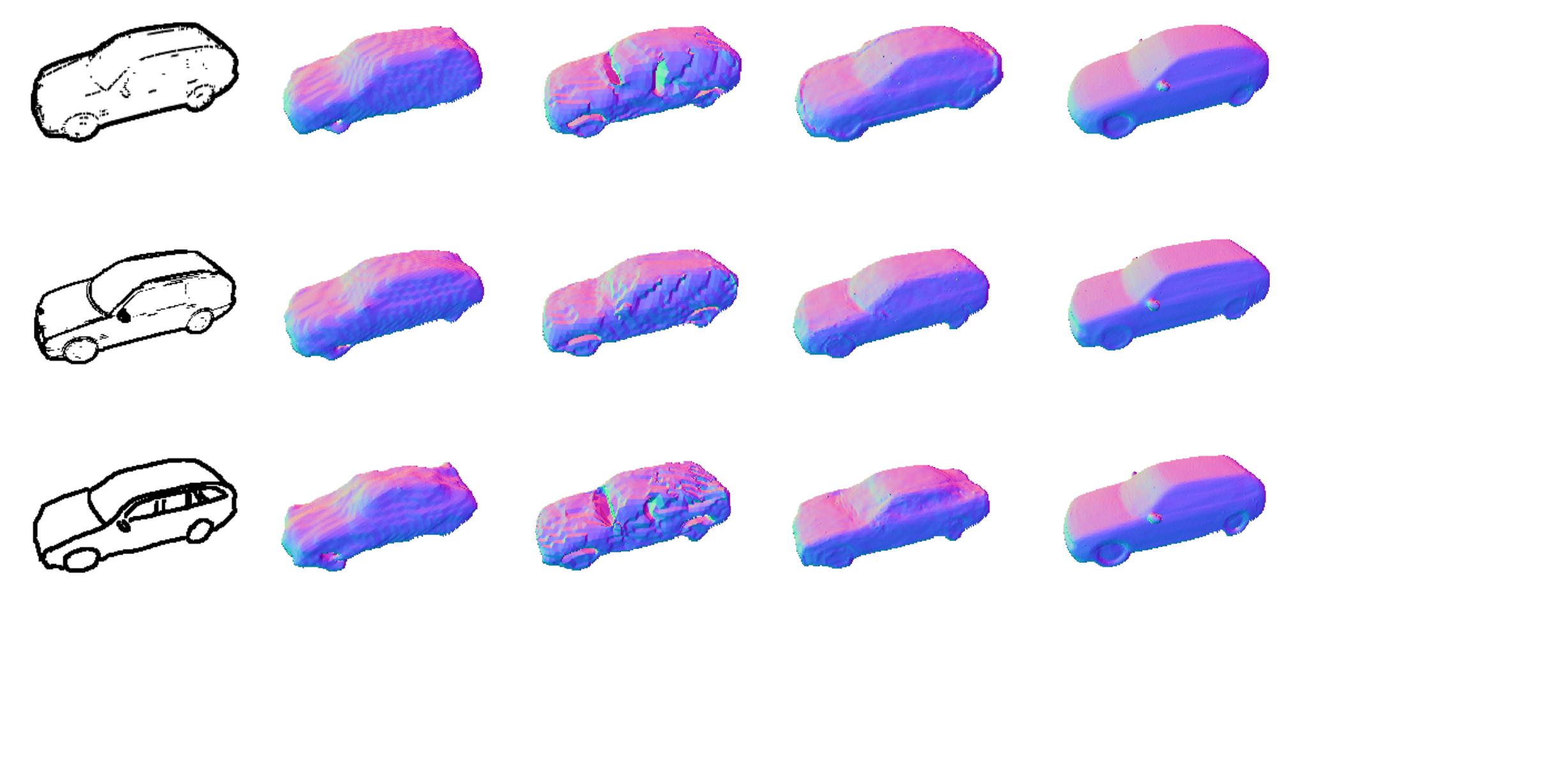}
 	\put(6,45){\tiny{input}}
 	\put(23,45){\tiny{Pix2Vox}~\cite{Delanoy18}}	
 	\put(41,45){\tiny{MeshRCNN}~\cite{Gkioxari19}}
 	\put(62,45){\tiny{DISN}~\cite{Xu19b}}
 	\put(80,45){\tiny{Sketch2Mesh}} 	
 \end{overpic}
\end{center}
\vspace{-3mm}
   \caption{\small \textbf{Robustness to changes in sketch style.} Given a \texttt{Suggestive} sketch (top), a \texttt{SketchFD}  one (middle), or a hand-drawn one (bottom), Sketch2Mesh---unlike  Pix2Vox, MeshRCNN, and DISN---yields reconstructions that are similar to each other and close to the ground-truth.}  
\label{fig:robustness}
\end{figure}

\subsection{Datasets}

Publicly available large-scale line-drawings datasets with associated 3D models are rare.  We therefore test our approach on two datasets, one for chairs that is available~\cite{Zhong20} and another for cars that we created ourselves. To further test, and crucially, to train our approach, we used 3D models from the well-established  ShapeNet~\cite{Chang15} to render 2D sketches. 

\parag{Rendered Car and Chair Sketches.} 

We use the car and chair categories from ShapeNet~\cite{Chang15} both for training and testing. We adopt the same train/test splits as in~\cite{Remelli20b}. For cars we use 1311 training samples and 113 test samples. The equivalent numbers are 5268 and 127 for chairs.  For each object and corresponding 3D mesh, we randomly sample 16 azimuth and elevation angles. The cameras point at the object centroid while their distance to it and their focal lengths are kept fixed. To demonstrate robustness to sketching style, we generate two different $256\times256$ binary sketches  for each viewpoint, as shown in the top two rows a Fig.~\ref{fig:robustness}. We will refer to them as \texttt{Suggestive} and \texttt{SketchFD} sketches, as described below. 


\sparag{Suggestive.}

We use the companion software of~\cite{DeCarlo03} to render sketches displaying that contain both occluding and suggestive contours. Suggestive contours are lines drawn on visible parts of the surface where a true occluding contour would first appear given a minimal viewpoint change. They are designed to emulate real line drawings in which lines other lines than the occluding contours are drawn to increase expressivity. 

\sparag{SketchFD.} 

We also use the older rendering approach of~\cite{Saito90}. We run an edge detector over the normal and depth maps of the rendered object. Edges in the depth map correspond to depth discontinuities while edges in the normal map correspond to sharp ridges and valleys. This yields synthetic sketches that, although conveying the same information, look very different from the ones on~\cite{DeCarlo03}, as can be seen on the left of Fig.~\ref{fig:robustness}.


\begin{figure}
\begin{center}
 	\begin{overpic}[clip, trim=0cm 7.0cm 0cm 3.2cm,width= .45\textwidth]{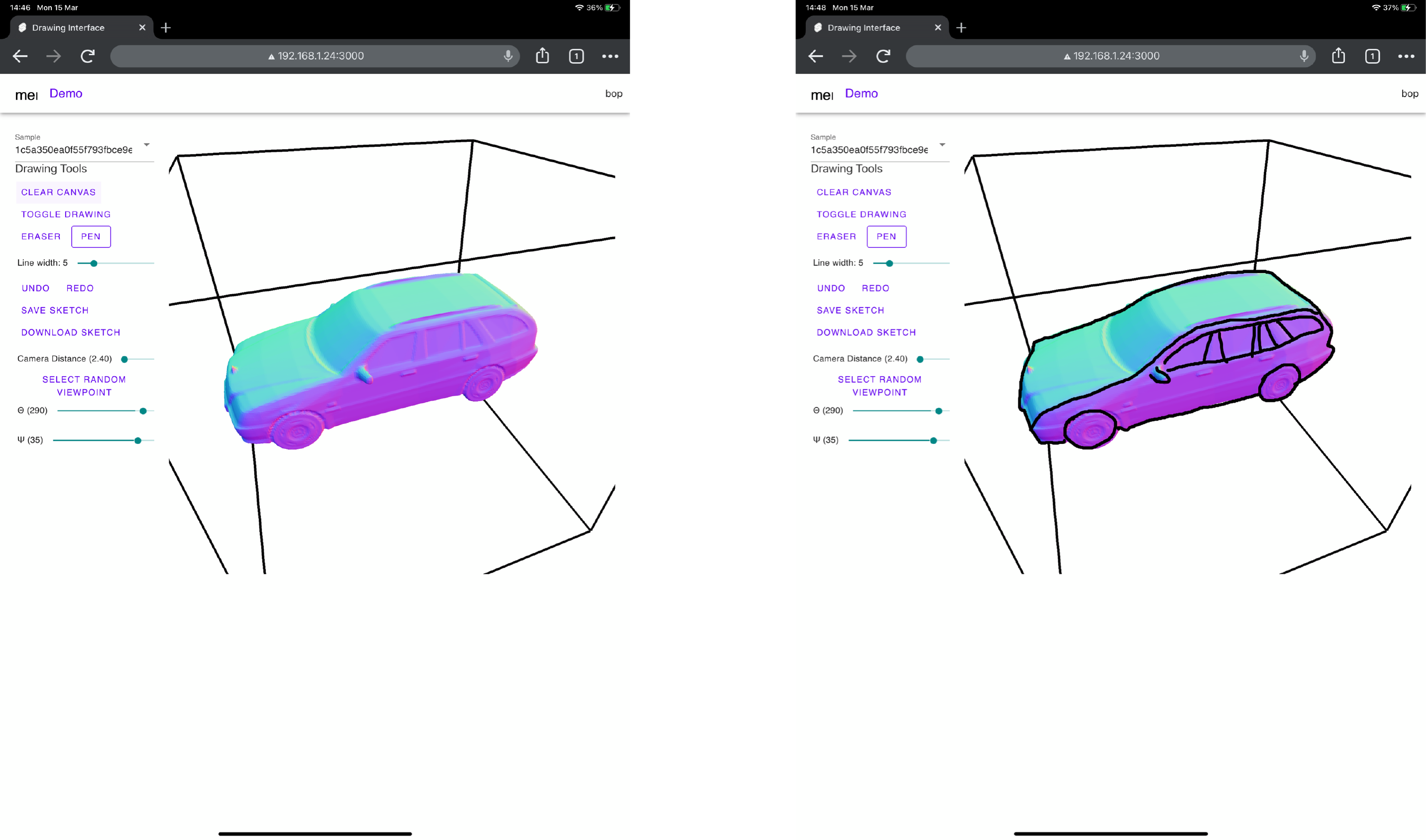}
 	\put(19,27){\small{(a)}}
 	\put(78,27){\small{(b)}}
 	
 	
 \end{overpic}
\end{center}
\vspace{-3mm}
   \caption{\small \textbf{Data acquisition interface.}  (a) To guide unexperienced users and limit imprecision, we display the normal map as seen from a specific viewpoint. (b) The user can use a pen to draw freely on the resulting image. }
\label{fig:data}\label{fig:tablet}
\end{figure}

\parag{Hand-Drawn Car Sketches}

We asked 5 students with no prior experience in 3D design to draw by-hand the 113 cars from the ShapeNet test set. To this end, we developed the sketching interface depicted by Fig.~\ref{fig:tablet} that runs on a standard tablet. The participants drew over normal maps rendered from the selected viewpoint so as to provide them guidance and ensure they all drew a similar car and used a known perspective. However, they were free to make the pen strokes they wanted. Hence, this dataset thus exhibits natural variations of style. To allow for comparison with results on the rendered sketches, we used the same viewpoint, which we will use to demonstrate that style change by itself is an obstacle to generalization for many methods.


\paragraph{Hand-Drawn Chair Sketches}


\begin{figure}
	\begin{center}
	 \includegraphics[width=.45\textwidth]{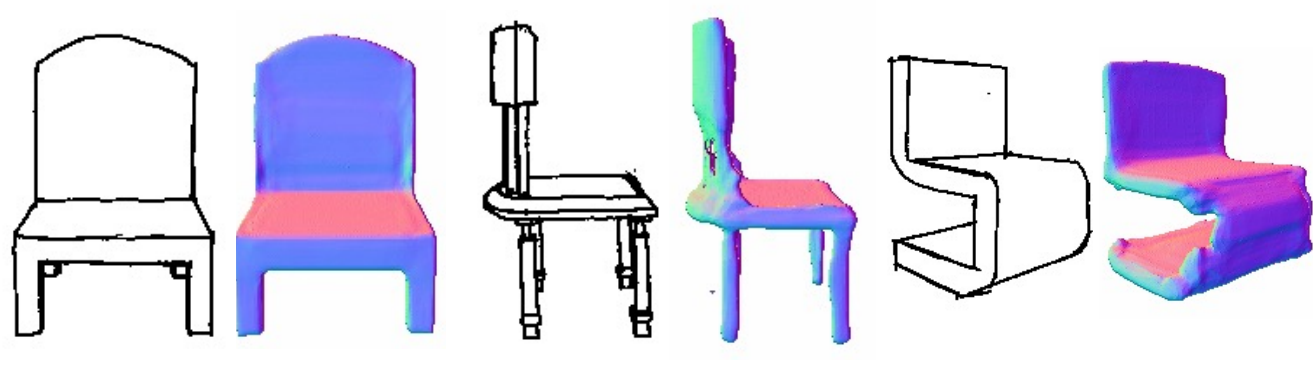}
	\end{center}
	\vspace{-3mm}
		 \caption{\textbf{ProSketch.} Input hand-drawn chair sketches~\cite{Zhong20} and \skm{} reconstructions.}
	\label{fig:prosketch}
\end{figure}

\comment{
\begin{figure*}
\begin{center}
 	\begin{overpic}[clip, trim=0cm 7.0cm 0cm 0cm,width= .95\textwidth]{figs/prosketch.pdf}

 	\put(5,36){\small{input}}
 	\put(18,36){\small{reconstruction}}
 	
 	\put(33,36){\small{silhouette gap}}
 	\put(49,36){\small{silhouette refinement}}
 	
 	\put(69,36){\small{contour gap}}
 	\put(85,36){\small{contour refinement}}
 	
 \end{overpic}
\end{center}
\vspace{-3mm}
   \caption{\textbf{reconstructions on ProSketch.} \ER{Compare all methods on this, harder because different cameras, different style...}. }
\label{fig:prosketch}
\end{figure*}
}

We use 177 chair sketches from the ProSketch dataset~\cite{Zhong20}. The chairs are seen from the  front, profile, or a 45\textdegree azimuth view, as shown in Fig.~\ref{fig:prosketch}. These viewpoints do not match the randomly selected ones we used for training, which makes this dataset especially challenging. Sample sketches and reconstructions are shown in Fig.~\ref{fig:prosketch}


\begin{table*}[ht]
	\begin{center}
	\begin{tabular}{cc}
		\scalebox{0.65}{
			\begin{tabular}{c|c|c|c|c}
				\Xhline{2\arrayrulewidth}
				Metric & Method & \multicolumn{3}{c}{Test Drawing Style} \\
				\cline{3-5}
				&  & Suggestive & SketchFD  & Hand-drawn \\
				\Xhline{2\arrayrulewidth}
				\multirow{3}{*}{CD-$l_2 \cdot 10^3$ $\downarrow$}
				&{\it Initial}       &1.613&4.658&6.818\\
				&\skmr &\textbf{1.400}&4.253&5.752\\
				&\skmc &1.420&\textbf{3.132}&\textbf{4.395}\\
				\cline{2-5}
				\Xhline{2\arrayrulewidth}
				\multirow{3}{*}{Normal Consistency $\uparrow$}
				&{\it Initial}        &91.14&84.73&81.40\\
				&\skmr &\textbf{92.41}&86.18&83.88\\
				&\skmc &92.20&\textbf{87.00}&\textbf{84.75}\\
				\cline{2-5}
					\Xhline{2\arrayrulewidth}
		\end{tabular}}
		&
		\scalebox{0.65}{
			\begin{tabular}{c|c|c|c|c}
				\Xhline{2\arrayrulewidth}
				Metric & Method & \multicolumn{3}{c}{Test Drawing Style} \\
				\cline{3-5}
				&  &  Suggestive & SketchFD  & Hand-drawn \\
				\Xhline{2\arrayrulewidth}
				\multirow{3}{*}{CD-$l_2 \cdot 10^3$ $\downarrow$}
				&{\it Initial}         &8.572&15.691&18.752\\
				&\skmr  &7.471&12.865&17.519\\
				&\skmc &\textbf{7.180}&\textbf{12.248}&\textbf{13.787}\\
				\cline{2-5}
					\Xhline{2\arrayrulewidth}
				
				\multirow{3}{*}{Normal Consistency $\uparrow$}
				&{\it Initial}        &80.86&72.83&61.17\\
				&\skmr &\textbf{83.99}&75.37&65.23\\
				&\skmc &{82.61}&\textbf{76.27}&\textbf{67.67}\\
				\cline{2-5}
					\Xhline{2\arrayrulewidth}
		\end{tabular}}
        \end{tabular}\\
        Cars \hspace{4cm}  \hspace{4cm} Chairs
	\end{center}
\caption{\label{tab:refinement} \small \textbf{Cars and Chairs.}
	Reconstruction metrics when using the encoding/decoding network trained on \texttt{Suggestive} synthetic sketches of cars and of chairs, and tested on all 3 datasets. We show \textit{initial} results before refinement and then using our two refinement methods. Note that \skmc{} does better than \skmr{}  on the styles it has {\it not} been trained for, indicating a greater robustness to style changes. 
}
\end{table*}


\begin{figure*}
\begin{center}
 	\begin{overpic}[clip, trim=0cm 2.5cm 0cm 0cm,width= .95\textwidth]{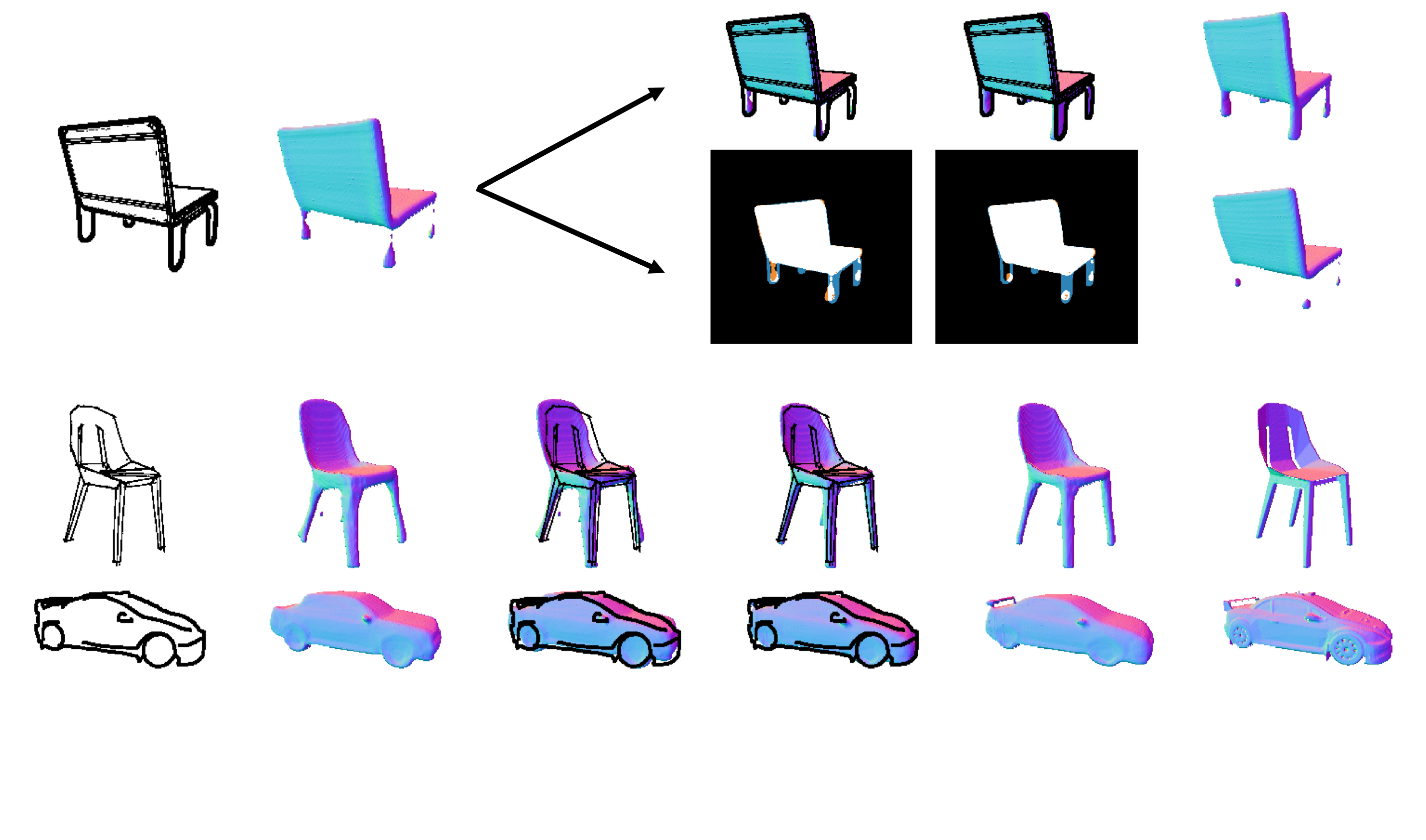}

 	\put(7,0){\small{input}}
 	\put(19,0){\small{reconstruction}}
 	
 	\put(39,0){\small{iter = 0}}
 	\put(55,0){\small{iter = 250}}
 	
 	\put(72,0){\small{refined}}
 	\put(87,0){\small{ground truth}}
 	
 	\put(32,28){\small{\skmr}}
 	\put(32,46){\small{\skmc}}
 	
 	\put(-2,12){{(b)}}
 	\put(-2,38){{(a)}}
 	
 	\put(7,44){\small{input}}
 	\put(19,44){\small{reconstruction}}
 	
 	\put(53,51){\small{iter = 0}}
 	\put(68,51){\small{iter = 250}}
 	
 	\put(85,51){\small{refined}}
 	
 \end{overpic}
\end{center}
\vspace{-1mm}
   \caption{\small \textbf{Mesh refinement.} (a) Comparison of \skmc{} (top) and \skmr{}  (bottom). \skmc{}  handles thin components such as the legs of the chair better because it leverages sparse information. We examine this in more detail in the Supplementary material. (b) \skmc{} results on challenging line drawings of a chairs and a car. We show the iterations from the initial mesh produced by the network that takes the sketch as input, which is then progressively refined. }
\label{fig:refinement}
\end{figure*}

\subsection{Metrics}

As reconstruction metric, we use a 3D Chamfer loss (CD-$l_2$, the lower the better). It is computed by sampling $N=10000$ points on the reconstructed mesh to form a first point cloud $\bC_1$ and $N$ on the ground truth mesh to form a second point cloud $\bC_2$. We then compute
\begin{small}
  \[
    \textrm{CD-}l_2 = \tfrac{1}{N} \sum_{x \in \bC_1} \min_{y \in \mathbf{y}}\left \| x-y \right \|^2 + \tfrac{1}{N} \sum_{y \in \bC_2} \min_{x \in \mathbf{x}}\left \| y-x \right \|^2 \; .
  \]
\end{small}

We also report  a normal consistency measure (NC, the higher the better), by taking the average pixel-wise dot product between normal maps of the reconstructed shape and the ground truth one.

\subsection{Choosing the Best Method} \label{subsec:choosing_best_method}

Recall from the method section, that we have proposed two variants of our approach to refining our 3D meshes. \skmr{} operates by turning the sketch into a foreground/background image and minimizing the distance between that image and the mesh projection while  \skmc{} deforms the mesh to minimize the 2D Chamfer distance between the external contours of its projection and those of the sketch. 

Once the latent representation has been learned on either  \texttt{Suggestive} or \texttt{SketchFD} contours, \skmc{}  can be used without any further training. By contrast,  \skmr{} requires an image translation network  to predict foreground/background masks from sketches. Here, we use the one of~\cite{Isola17} with a UNet~\cite{Ronneberger15} as its generator and in the LSGAN setting~\cite{Mao16}. We train four separate instances of it on ShapeNet , one for each shape category (cars and chairs) and for each sketch rendering style (\texttt{Suggestive} and \texttt{SketchFD}). 

This being done, we can compare \skmr{} against \skmc{} on the test sets for both categories of object and the three categories of drawing we use, \texttt{Suggestive}, \texttt{SketchFD}, and \texttt{Hand Drawn}. We show qualitative results in Figs.~\ref{fig:prosketch} and~\ref{fig:refinement}. We report quantitative results in Tab.~\ref{tab:refinement} for models trained on \texttt{Suggestive} contours. Similar results on \texttt{SketchFD} contours are presented in the supplementary material. Overall, both \skmr{} and \skmc{} improve the initial metrics but \skmc{} appears to be more robust to style changes. In other words, 
\skmr{} overfits to the style it is trained on and does not do as well as \skmc{} when tested on a different one. Adding this to the fact, that \skmc{}, unlike \skmr{}, does not require to train an auxiliary network clearly makes it the better approach. We will therefore use it in the remainder of the paper except otherwise noted and will refer to it as \skm{} for brevity.


\begin{table}[ht]
    \caption{\textbf{Comparative results on Cars.}   }
	\label{tab:robustness_cars_decarlo}
	\begin{center}
		\scalebox{0.65}{
			\begin{tabular}{c|c|c| c | c}
				\Xhline{2\arrayrulewidth}
				\Xhline{2\arrayrulewidth}
				\multicolumn{5}{c}{\textbf{Training Drawing Style: Suggestive}} \\
				\Xhline{2\arrayrulewidth}
				Metric & Method & \multicolumn{3}{c}{Test Drawing Style} \\
				\cline{3-5}
				&  & Suggestive& SketchFD  & Hand-drawn \\
				\Xhline{2\arrayrulewidth}
				
				\multirow{4}{*}{CD-$l^2 \cdot 10^3$ $\downarrow$}
				&Pix2Vox~\cite{Delanoy18}&2.336&6.237&8.599\\
				&MeshRCNN~\cite{Gkioxari19}&3.491&6.923&7.849\\
				&DISN~\cite{Xu19b}&1.529&7.764&10.396\\
				\cline{2-5}
				&Sketch2Mesh &\textbf{1.420}&\textbf{3.132}&\textbf{4.396}\\
				\Xhline{2\arrayrulewidth}
				
				\multirow{4}{*}{Normal Consistency $\uparrow$}
				&Pix2Vox~\cite{Delanoy18}&89.07&80.49&76.70\\
				&MeshRCNN~\cite{Gkioxari19}&84.19&79.93&77.91\\
				&DISN~\cite{Xu19b}&{92.15}&79.51&72.52\\
				\cline{2-5}
				&Sketch2Mesh  &\textbf{92.20}&\textbf{87.00}&\textbf{84.74}\\
				\Xhline{2\arrayrulewidth}
				\Xhline{2\arrayrulewidth}
				\multicolumn{5}{c}{\textbf{Training Drawing Style: SketchFD}} \\
				\Xhline{2\arrayrulewidth}
				
				\multirow{4}{*}{CD-$l^2 \cdot 10^3$ $\downarrow$}
                &Pix2Vox~\cite{Groueix18a}&3.529&2.475&3.146\\
                &MeshRCNN~\cite{Gkioxari19}&3.117&3.596&4.829\\
                &DISN~\cite{Xu19b}&4.036&{1.573}&3.763\\
				\cline{2-5}
				&Sketch2Mesh  &\textbf{2.419}&\textbf{1.516}&\textbf{2.047}\\
				
				\Xhline{2\arrayrulewidth}
				
				\multirow{4}{*}{Normal Consistency $\uparrow$}
				&Pix2Vox~\cite{Groueix18a}&87.11&89.21&86.27\\
				&MeshRCNN~\cite{Gkioxari19}&83.22&82.81&80.83\\
                &DISN~\cite{Xu19b}&86.34&91.30&87.66\\
				\cline{2-5}
				&Sketch2Mesh  &\textbf{91.23}&\textbf{92.09}&\textbf{91.03}\\
				\Xhline{2\arrayrulewidth}
				\Xhline{2\arrayrulewidth}
				
		\end{tabular}}
	\end{center}
\end{table}

\comment{\begin{table}[ht]
		\caption{\textbf{SVR on ShapeNet/Cars.} \ER{moved out DISN silhouette, makes more sense for experience 2.4 + let's show that SVR methods fail! }  }
		\label{tab:robustness_cars_DR}
		\begin{center}
			\scalebox{0.65}{
				\begin{tabular}{c|c|c| c | c}
					\Xhline{2\arrayrulewidth}
					\multicolumn{5}{c}{\textbf{Training Drawing Style: SketchDR}} \\
					\Xhline{2\arrayrulewidth}
					Metric & Method & \multicolumn{3}{c}{Test Drawing Style} \\
					\cline{3-5}
					&  & SketchDR & DeCarlo & Human \\
					\Xhline{2\arrayrulewidth}
					\multirow{7}{*}{CD-$l^2 \cdot 10^3$ $\downarrow$}
					&Pix2Vox~\cite{Groueix18a}&2.475&3.529&3.146\\
					\cline{2-5}
					&MeshRCNN~\cite{Gkioxari19}&3.220&3.186&4.346\\
					&DISN~\cite{Groueix18a}&1.773&4.036&3.763\\
					&UCLID-Net~\cite{Guillard20}&1.306&2.034&1.553\\
					\cline{2-5}
					&MeshSDF  &1.815&3.227&2.534\\
					&AtlasNet~\cite{Groueix18a}&1.636&2.386&2.057\\
					\Xhline{2\arrayrulewidth}
					
					\multirow{7}{*}{Normal Consistency $\uparrow$}
					&Pix2Vox~\cite{Groueix18a}&89.21&87.11&86.27\\
					\cline{2-5}
					&MeshRCNN~\cite{Gkioxari19}&81.24&81.08&78.91\\
					&DISN~\cite{Groueix18a}&90.30&86.34&87.66\\
					&UCLID-Net~\cite{Guillard20}&48.27&44.91&48.02\\
					\cline{2-5}
					&MeshSDF  &90.94&89.67&89.06\\
					&AtlasNet ~\cite{Groueix18a}&87.62&87.28&85.95\\
					\Xhline{2\arrayrulewidth}
					\multicolumn{5}{c}{\textbf{Training Drawing Style: DeCarlo}} \\
					\Xhline{2\arrayrulewidth}
					
					\multirow{6}{*}{CD-$l^2 \cdot 10^3$ $\downarrow$}
					&Pix2Vox~\cite{Groueix18a}&6.237&2.336&8.599\\
					\cline{2-5}
					&MeshRCNN~\cite{Gkioxari19}&7.157&3.902&8.200\\
					&DISN~\cite{Groueix18a}&7.764&1.529&10.396\\
					&UCLID-Net~\cite{Guillard20}&5.271&1.325&6.813\\
					\cline{2-5}
					&MeshSDF  &4.643&1.612&6.841\\
					&AtlasNet ~\cite{Groueix18a}&3.552&1.532&4.592\\
					\Xhline{2\arrayrulewidth}
					
					\multirow{6}{*}{Normal Consistency $\uparrow$}
					&Pix2Vox~\cite{Groueix18a}&80.49&89.07&76.70\\
					\cline{2-5}
					&MeshRCNN~\cite{Gkioxari19}&76.96&82.68&74.43\\
					&DISN~\cite{Groueix18a}&79.51&92.15&72.52\\
					&UCLID-Net~\cite{Guillard20}&49.03&50.93&47.73\\
					\cline{2-5}
					&MeshSDF  &84.77&91.21&81.44\\
					&AtlasNet ~\cite{Groueix18a}&83.46&87.63&81.50\\
					\Xhline{2\arrayrulewidth}
			\end{tabular}}
		\end{center}
\end{table}}

\begin{table}[ht]
    \caption{\textbf{Comparative results on Chairs.}  }
	\label{tab:chairs_sketchDR}
	\begin{center}
		\scalebox{0.65}{
			\begin{tabular}{c|c|c| c|c}
				\Xhline{2\arrayrulewidth}
				\Xhline{2\arrayrulewidth}
				\multicolumn{5}{c}{\textbf{Training Drawing Style: Suggestive}} \\
				\Xhline{2\arrayrulewidth}
				Metric & Method & \multicolumn{3}{c}{Test Drawing Style} \\
				\cline{3-5}
				&  & Suggestive & SketchFD & Hand-drawn \\
				\Xhline{2\arrayrulewidth}
				
				\multirow{4}{*}{CD-$l^2 \cdot 10^3$ $\downarrow$}
				&Pix2Vox~\cite{Delanoy18}&22.953&33.46&62.132\\
				&MeshRCNN~\cite{Gkioxari19}&\textbf{6.775}&\textbf{10.718}&19.055\\
				&DISN~\cite{Xu19b}&7.045&18.104&23.282\\
				\cline{2-5}
				&Sketch2Mesh &7.180&12.248&\textbf{13.787}\\
				\Xhline{2\arrayrulewidth}
				
				\multirow{4}{*}{Normal Consistency $\uparrow$}
				&Pix2Vox~\cite{Groueix18a}&73.01&64.28&40.12\\
				&MeshRCNN~\cite{Gkioxari19}&76.91&72.77&58.03\\
				&DISN~\cite{Xu19b}&80.44&54.10&51.81\\
				\cline{2-5}
				&Sketch2Mesh &\bf{82.61}&\textbf{76.27}&\textbf{67.67}\\
				
				\Xhline{2\arrayrulewidth}
				\Xhline{2\arrayrulewidth}
				\multicolumn{5}{c}{\textbf{Training Drawing Style: SketchFD}} \\
				\Xhline{2\arrayrulewidth}
				
				\multirow{4}{*}{CD-$l^2 \cdot 10^3$ $\downarrow$}
                &Pix2Vox~\cite{Groueix18a}&34.759&22.690&46.687\\
                &MeshRCNN~\cite{Gkioxari19}&\textbf{9.530}&\textbf{5.812}&16.620\\
                &DISN~\cite{Xu19b}&13.059&8.628&18.104\\
				\cline{2-5}
				&Sketch2Mesh &\textbf{9.524}&6.737&\textbf{12.585}\\
				\Xhline{2\arrayrulewidth}
				
				\multirow{4}{*}{Normal Consistency $\uparrow$}
				&Pix2Vox~\cite{Groueix18a}&65.97&72.52&52.96\\
				&MeshRCNN~\cite{Gkioxari19}&77.62&\textbf{84.75}&69.76\\
                &DISN~\cite{Xu19b}&73.39&80.21&62.58\\
				\cline{2-5}
				&Sketch2Mesh &\textbf{81.00}&83.10&\textbf{70.39}\\
				\Xhline{2\arrayrulewidth}
				\Xhline{2\arrayrulewidth}
				
		\end{tabular}}
	\end{center}
\end{table}

\subsection{Comparison against State-of-the-Art Methods}

We now compare \skm{} against state-of-the-art methods that produce watertight meshes as we do. To this end, we train the architecture of~\cite{Delanoy18} that regresses volumetric grids from sketches, which we dub {\it Pix2Vox}. We also compare to recent SVR method that rely on perceptual feature pooling from the image plane DISN~\cite{Xu19b} and MeshRCNN~\cite{Gkioxari19}. For a fair comparison, we use them in conjunction with the same image encoder as we do, ResNet18~\cite{He16a}. 


\begin{figure}
\begin{center}
\begin{tabular}{cccc}
 \includegraphics[width=.1\textwidth]{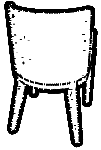}&
 \includegraphics[width=.1\textwidth]{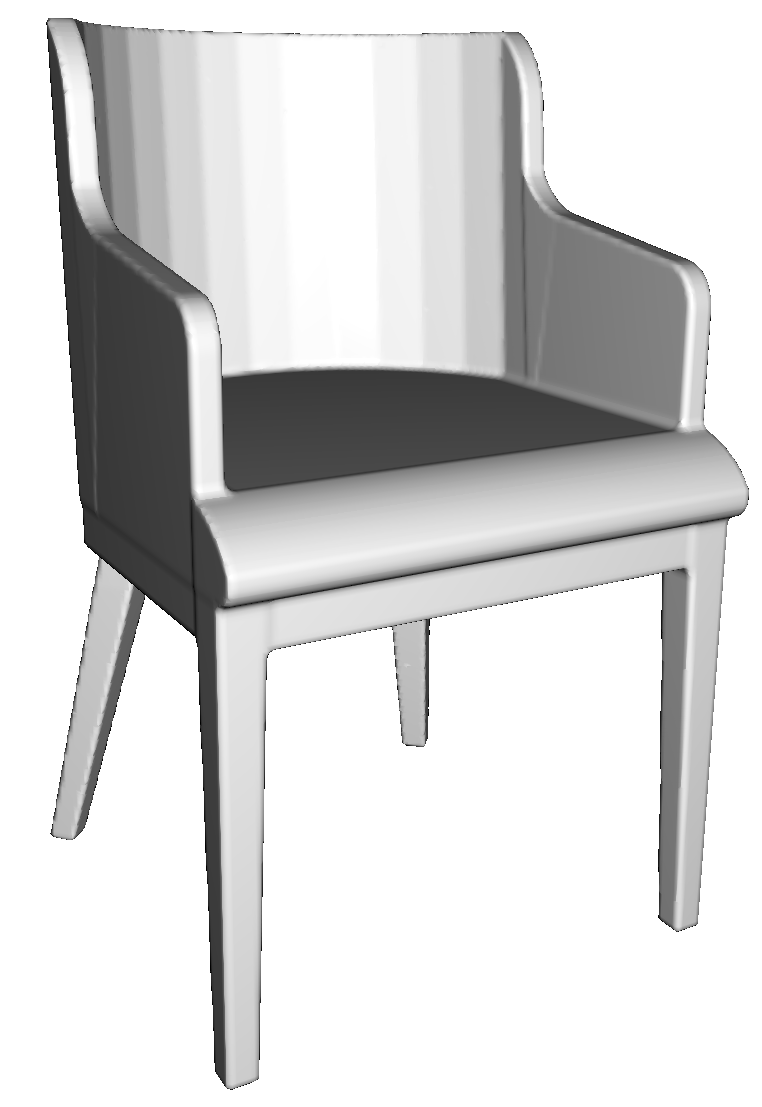} &
 \includegraphics[width=.1\textwidth]{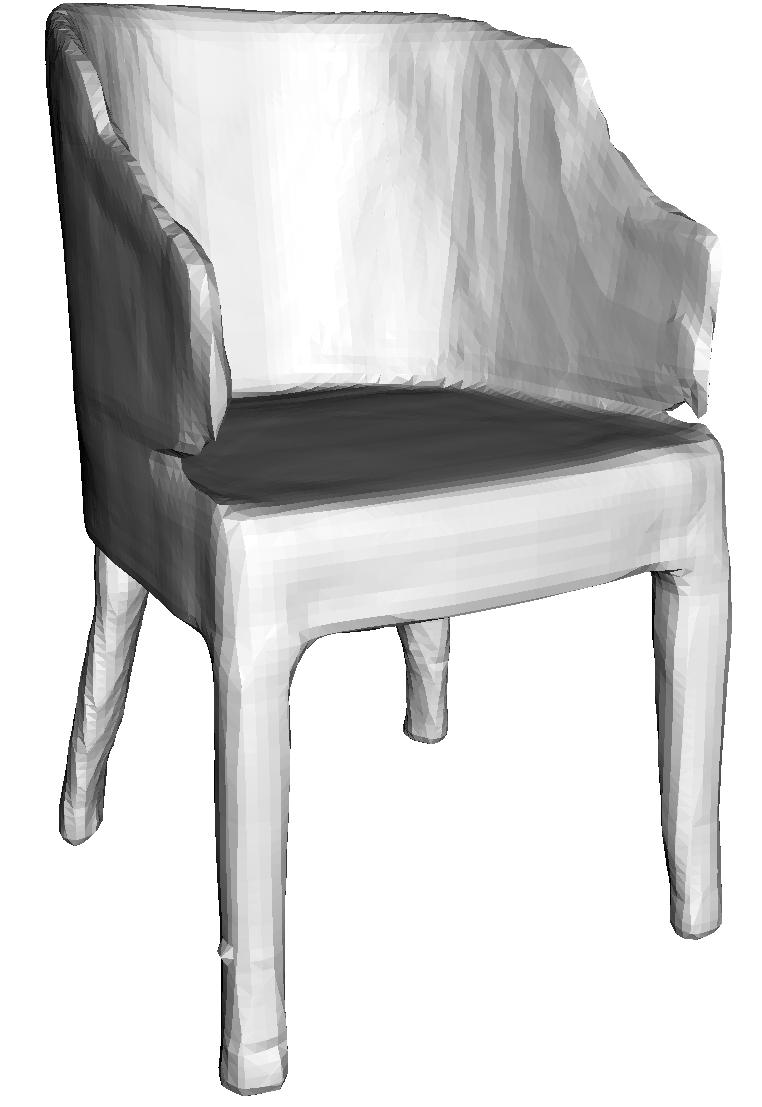} &
 \includegraphics[width=.1\textwidth]{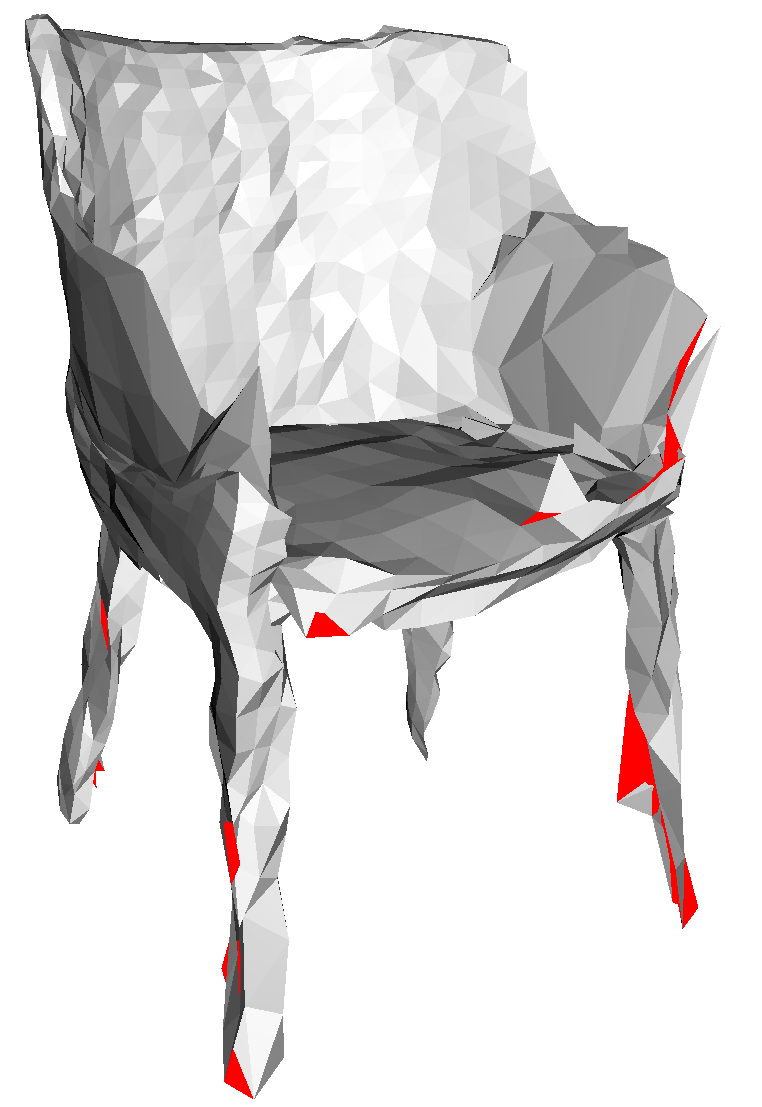} \\
    (a) & (b) & (c) & (d)
\end{tabular}
\end{center}
\vspace{-3mm}
   \caption{\textbf{Comparison with MeshRCNN:} \textbf{(a)} Input sketch \textbf{(b)} Ground truth shape, \textbf{(c)} \textit{Sketch2Mesh} reconstruction, with CD-$l_2$=1.98, \textbf{(d)} MeshRCNN reconstruction, with CD-$l_2$=1.91. The flipped facets are shown in red. Despite having a slightly higher CD-$l_2$, our reconstruction is far more usable for further processing and, arguably, resembles the ground truth more than the MeshRCNN one.}
\label{fig:meshrcnn_comparison}
\end{figure}

We show qualitative results in Fig.~\ref{fig:robustness}. We report quantitative results on ShapeNet Cars and Chairs in Tables~\ref{tab:robustness_cars_decarlo} and~\ref{tab:chairs_sketchDR} when the latent representation have been learned either on \texttt{Suggestive} or \texttt{SketchFD} contours. On cars, \skm{} clearly outperforms the other methods. On chairs, MeshRCNN is very competitive, especially in terms of CD-$l_2$. But, as shown in Fig.~\ref{fig:meshrcnn_comparison}, the meshes it produces are hardly usable, even though we uses the \textit{Pretty} setup of the algorithm that attempts to regularize them. This is a well known phenomenon reported by its authors themselves. By contrast, our meshes can directly be used for downstream applications, without further preprocessing.


\begin{table}[ht]
    \caption{\small \textbf{Comparison with the approach of~\cite{Song20b}.}}
	\label{tab:silho_conditioning}
	\begin{center}
		\scalebox{0.85}{
			\begin{tabular}{c|c| c}
				\Xhline{2\arrayrulewidth}
				\Xhline{2\arrayrulewidth}
				 Method & \multicolumn{2}{c}{Metric} \\
				\cline{2-3}
				 & CD-$l^2 \cdot 10^3$ $\downarrow$ & NC $\uparrow$ \\
				\Xhline{2\arrayrulewidth}
				MeshSDF~\cite{Remelli20b}    &3.231&89.67\\
				\cline{1-3}
				MeshSDF~\cite{Remelli20b} + mask  &3.124&90.05\\
				\cline{1-3}
				\skmr{}  &2.538&90.92\\
				\cline{1-3}
				\skmc{} & {\bf 2.419}&{\bf 91.23}\\
				\Xhline{2\arrayrulewidth}
				\Xhline{2\arrayrulewidth}
		\end{tabular}}
	\end{center}
\end{table}

\comment{
\begin{table}[ht]
    \caption{\textbf{Comparison of conditioning vs. refining with foreground/background mask} We train MeshSDF networks on {\it SketchFD} cars, and test them on {\it Suggestive}. Four approaches are compared: directly using the standard approach ({\it raw}), feeding the foreground/background mask as additional input to the network ({\it raw} + mask), doing refinement with the mask (+ $\mathcal{L}_{F/B}$) or with external contours matching (+ $\mathcal{L}_{CD}$). Masks can either be ground truth (GT) or estimated by a pix2pix (pred).}
	\label{tab:silho_conditioning}
	\begin{center}
		\scalebox{0.85}{
			\begin{tabular}{c|c| c}
				\Xhline{2\arrayrulewidth}
				\Xhline{2\arrayrulewidth}
				 Method & \multicolumn{2}{c}{Metric} \\
				\cline{2-3}
				 & CD-$l^2 \cdot 10^3$ $\downarrow$ & NC $\uparrow$ \\
				\Xhline{2\arrayrulewidth}
				MeshSDF \textit{raw}    &3.231&89.67\\
				\cline{1-3}
				MeshSDF \textit{raw} + (GT mask)  &2.602&90.45\\
				MeshSDF \textit{raw}  + (pred mask ) &3.124&90.05\\
				\cline{1-3}
				MeshSDF + $\mathcal{L}_{F/B}$ (GT mask)  &2.260&91.06\\
				MeshSDF + $\mathcal{L}_{F/B}$ (pred mask ) &2.538&90.92\\
				\cline{1-3}
				MeshSDF + $\mathcal{L}_{CD}$ &2.419&91.23\\
				\Xhline{2\arrayrulewidth}
				\Xhline{2\arrayrulewidth}

		\end{tabular}}
	\end{center}
\end{table}
}

\comment{
\begin{table}[ht]
	\caption{\textbf{Using binary mask information on ShapeNet/Cars/Ours.} }
	\label{tab:silho_conditioning2}
	\begin{center}
		\scalebox{0.85}{
			\begin{tabular}{c|c| c}
				
				\Xhline{2\arrayrulewidth}
				Method & \multicolumn{2}{c}{Metric} \\
				\cline{2-3}
				& CD-$l^2 \cdot 10^3$ $\downarrow$ & NC $\uparrow$ \\
				\Xhline{2\arrayrulewidth}
				MeshSDF  &2.534&89.06\\
				\cline{1-3}
				MeshSDF (+ GT silho)  &2.368&90.19\\
				MeshSDF (+ pred silho ) &2.433&89.96\\
				\cline{1-3}
				MeshSDF (ref GT silho)  &2.005&91.15\\
				MeshSDF (ref pred silho ) &2.054&91.02\\
				\cline{1-3}
				MeshSDF (chamfer ) &2.047&91.03\\
				\Xhline{2\arrayrulewidth}

		\end{tabular}}
	\end{center}
\end{table}
}

\begin{figure*}[t!]
	
	\begin{center}
		\begin{overpic}[clip, trim=0cm 7.0cm 3cm 0 cm,width= .95\textwidth]{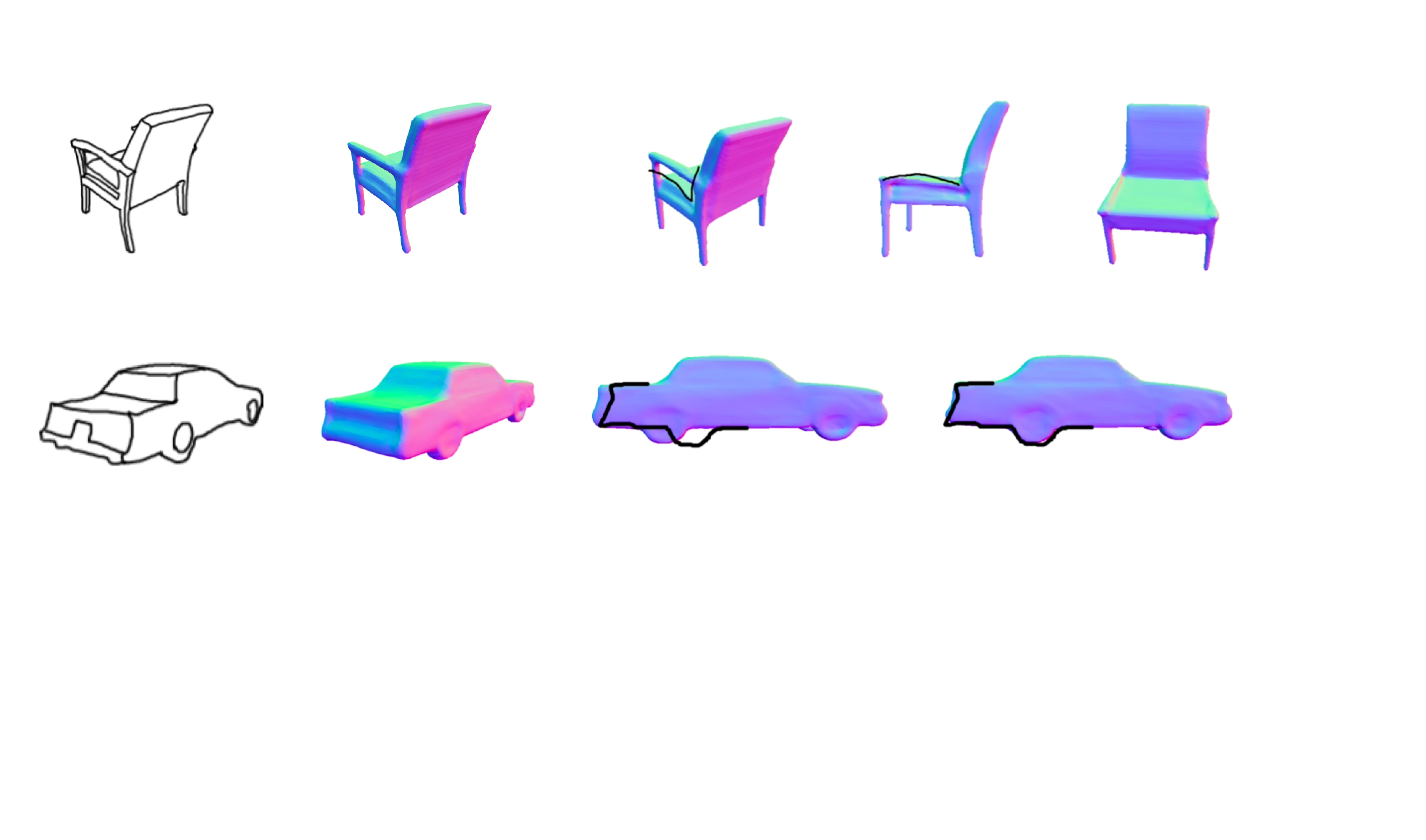}
			\put(4,35){{(a) Input sketch}}
			\put(25,35){{(b) Reconstructing}}
			\put(70,35){{(c) Editing}}
		
		\end{overpic}
	\end{center}

\vspace{-3mm}
   \caption{\small \textbf{Interactive reconstruction \& editing.} 
   	We developed an interface where the user can draw an initial sketch (a) to obtain its 3D reconstruction (b). It can then manipulate the object in 3D and draw one or more desired modifications (c). 3D surfaces are then optimized to match each constraint, solving the optimization problem of Section \ref{sec:partial}. The strong prior learned by our model allows to preserve global properties such as symmetry despite users provide sparse 2D strokes in input. Best seen in supplemental video.}
\label{fig:partial_sketch}
\end{figure*}

For completeness, we note that a very recent paper~\cite{Song20b} also advocates using foreground/background masks to improve 3D reconstruction from sketches. However, instead of refining the mesh produced by a network using such as a mesh as done by \skmr{}, it recommends feeding the mask as an additional input to the network that produces the initial 3D shape. In Tab.~\ref{tab:silho_conditioning}, we compare this approach to ours when the network is trained using the \texttt{SketchFD} sketches on cars and tested on \texttt{Suggestive}. Both \skmr{} and  \skmc{} outperform it.  


\subsection{Interactive 3D editing}

An important feature of \skm{} is that is can exploit sketches made of a single stroke to refine previously obtained shapes as discussed in Section~\ref{sec:partial}, as shown in Fig.~\ref{fig:partial_sketch}. To showcase the interactivity of our approach we built a web based user interface. The user may draw a sketch with the mouse or a touch enabled device and submit it to \skm{}. Then, successive partial sketches can also be input and matched by the optimizer. A video is provided in the supplementary material to show it in action.

\section{Conclusion}

We have proposed an approach to deriving 3D shapes from sketches that relies on an encoder/decoder architecture to compute a latent surface representation of the sketch. It can in turn be refined to match the external contours of the sketch. It handles sketches drawn in a style it was not specifically trained for and outperforms state-of-the-art methods. Furthermore, it allows for interactive refinements by specifying partial 2D contours the object's projection must match, provided that perspective camera parameters are associated to the sketch. This can be achieved easily on a tablet using a stylus-based interface to draw.

We can see two natural improvements to our work. One is linked to the learned priors in our parametrization. Although the priors are usually good at preserving global shape properties such as symmetry, they can be either too constraining or not enough when for partial refinements. We would like some priors to actually be constraints---the wheels of the cars must be round and cannot touch the wheel wells and the feet of the chairs must all have the same length, for example---in addition to those imposed by 2D sketches so that our technique can be turned into a full-fledged tool for Computer Assisted Design.Another research direction would be to incorporate interior lines in our refinement process. This is also an interesting challenge since we don't want to sacrifice the generalization ability this simple technique allowed us to achieve. 

\section{Acknowledgments}

This work was supported in part by the Swiss National Science Foundation and by the Swiss Innovation Agency.


{\small
\bibliographystyle{ieee_fullname}
\bibliography{string,vision,graphics,learning,biomed,egbib}
}

\clearpage


\section{Supplementary Material}

\subsection{External Contours}
Our 2D Chamfer refinement objective for matching external contours requires an estimation of these contours. We here describe the simple algorithms we use to get them for the reconstructed mesh, and for the full input sketch.

\subsubsection{External Contours of Reconstructed Shapes}


\begin{figure}[ht]
\begin{center}
   \setlength{\fboxsep}{0pt}%
   \setlength{\fboxrule}{0.8pt}%
\begin{tabular}{cccc}
    \includegraphics[width=.097\textwidth]{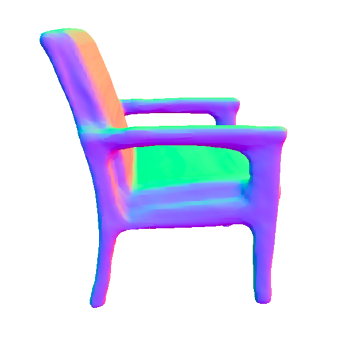}
    &
     \includegraphics[width=.097\textwidth]{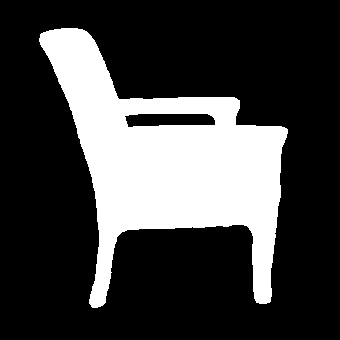}
    &
 \fbox{\includegraphics[width=.097\textwidth]{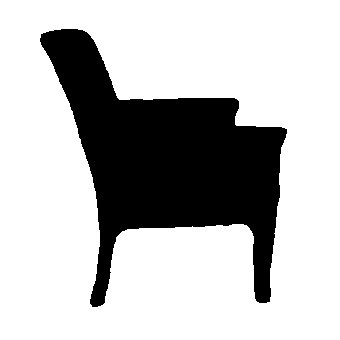}} &
    \includegraphics[width=.097\textwidth]{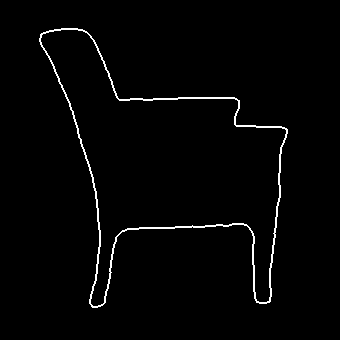}
     \\
    (a) & (b) & (c) & (d)
\end{tabular}
\end{center}
\vspace{-3mm}
   \caption{\textbf{External contours of reconstructed shape:} \textbf{(a)} Initially reconstructed shape \textbf{(b)} Rendered foreground/background mask, \textbf{(c)} Flood-filling (b) from one image corner, and performing 1 pixel dilation of the flood-filled background \textbf{(d)} Taking the pixel-wise multiplication of (b) and (c) yields an exterior contour image, in which internal holes are ignored (the armrest for example here).}
\label{fig:filter_contours_floodfilling}
\end{figure}

Given mesh $\cM_{\Theta}$ and projection $\Lambda$, we render a ${H\times W}$ foreground/background mask. Then we flood fill the background, starting from one image corner, and apply morphological dilation to the result. As depicted on Fig.~\ref{fig:filter_contours_floodfilling}, taking the pixel-wise multiplication of this dilated flood filled background with the original foreground/background mask yields an external contour that ignores inner holes. We use this image for $\widetilde{\bF}$ in Section~\ref{sec:contours}.

\subsubsection{External Contours of Input Sketches}


\begin{figure}[ht]
   \begin{center}
      \begin{overpic}[width=.47\textwidth]{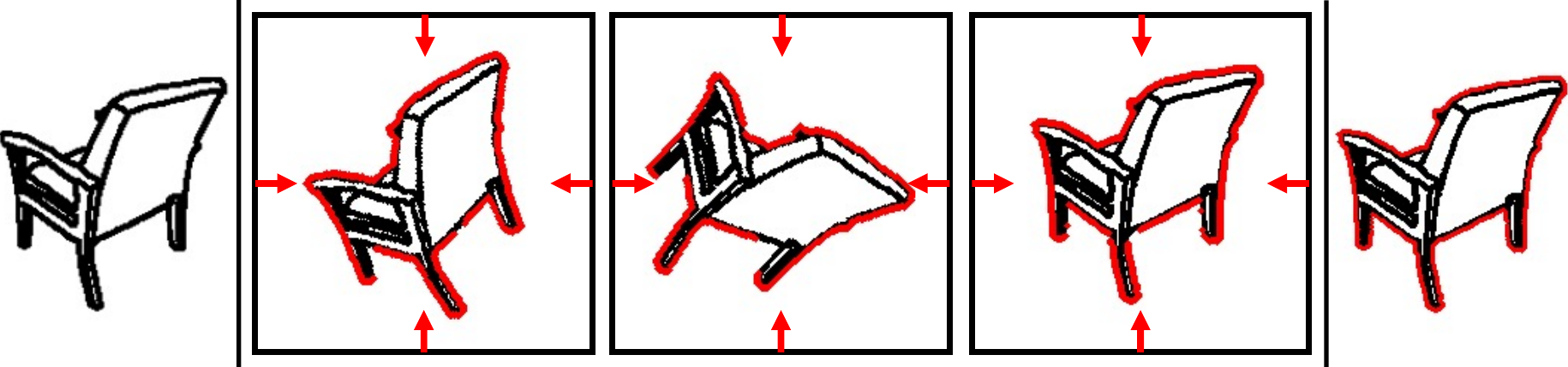} 
         \put(2,-3){\small{\textit{Input}}} 
         \put(40,-3){\small{\textit{Ray shooting}}} 
         \put(87,-3){\small{\textit{Output}}} 
      \end{overpic} 
   \end{center}
      \caption{\textbf{External contours of input sketch:} We rotate the input sketch at various angles and shoot vertical and horizontal rays to only keep the first encountered pen stroke (red pixels). These pixels obtained at different rotation angles are then aggregated to yield the full external contour (shown in red on the last panel, superposed to the sketch).}
   \label{fig:filter_contours_rays}
   \end{figure}

\comment{
\begin{figure}[ht]
\begin{center}
   \addtolength{\tabcolsep}{-1.2pt}    
\begin{tabular}{ccccc}
   \includegraphics[width=.08\textwidth]{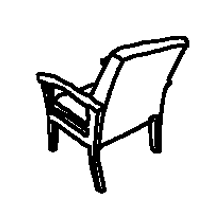} &
   \includegraphics[width=.08\textwidth]{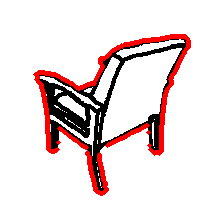} &
   \includegraphics[width=.08\textwidth]{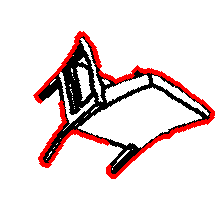} &
   \includegraphics[width=.08\textwidth]{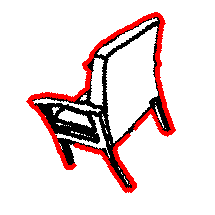} &
   \includegraphics[width=.08\textwidth]{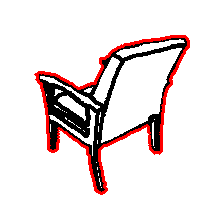} \\
    (a) & (b) & (c) & (d) & (e)
\end{tabular}
\addtolength{\tabcolsep}{1.2pt}    
\end{center}
\vspace{-3mm}
   \caption{\textbf{External contours of input sketch:} Given input sketch (a), we rotate it at various angles and shoot vertical and horizontal rays to only keep the first encountered pen stroke (red pixels in (b), (c), (d)). These pixels obtained at different rotation angles are then aggregated to yield the full external contour (e).}
\label{fig:filter_contours_rays}
\end{figure}
}
 
Given an input sketch, we cannot apply the above method since line drawings might not be watertight. Instead, we apply an image-space only algorithm that extracts external contours, which can then be matched against the ones of the initial reconstruction.

As pictured in Fig.~\ref{fig:method}(c), we propose to do this by shooting rays from the image borders, at multiple angles, and only preserve the first encountered stroke for each ray. For the ease of implementation, in practice we shoot rays that are perpendicular to the image borders, but rotate the input image of $\pm \left \{ 0, 10, 20, 30, 35, 40, 45  \right \}$ degrees and aggregate the resulting pixels at each angle. This is depicted in Fig.~\ref{fig:filter_contours_rays}.

To achieve a relative invariance to pen size (free choice in our interface), we extract both the entry and exit pixels of the first pen stroke a ray encounters. In case the average distance over the whole image between the entry and exit pixels is greater than a threshold, we heuristically consider the line as thick and only keep the exit pixels - this corresponds to the inner shell of the external contour. Otherwise, we consider the line as thin, and keep the entry pixel.

\subsection{Comparison of the two Refinement Approaches}

\subsubsection{Training on \texttt{SketchFD}}

\begin{table*}[th]
	\begin{center}
	\begin{tabular}{cc}
		\scalebox{0.65}{
			\begin{tabular}{c|c|c|c|c}
				\Xhline{2\arrayrulewidth}
				Metric & Method & \multicolumn{3}{c}{Test Drawing Style} \\
				\cline{3-5}
				&  & Suggestive & SketchFD  & Hand-drawn \\
				\Xhline{2\arrayrulewidth}
				\multirow{3}{*}{CD-$l_2 \cdot 10^3$ $\downarrow$}
				&{\it Initial}	&3.231					&1.815					&2.534\\
				&\skmr 					&2.538					&\textbf{1.515}	&2.054\\
				&\skmc 					&\textbf{2.419}	&1.516					&\textbf{2.047}\\
				\cline{2-5}
				\Xhline{2\arrayrulewidth}
				\multirow{3}{*}{Normal Consistency $\uparrow$}
				&{\it Initial}	&89.67					&90.94					&89.06\\
				&\skmr 					&90.92					&\textbf{92.34}	&91.02\\
				&\skmc 					&\textbf{91.23}	&92.09					&\textbf{91.03}\\
				\cline{2-5}
					\Xhline{2\arrayrulewidth}
		\end{tabular}}
		&
		\scalebox{0.65}{
			\begin{tabular}{c|c|c|c|c}
				\Xhline{2\arrayrulewidth}
				Metric & Method & \multicolumn{3}{c}{Test Drawing Style} \\
				\cline{3-5}
				&  &  Suggestive & SketchFD  & Hand-drawn \\
				\Xhline{2\arrayrulewidth}
				\multirow{3}{*}{CD-$l_2 \cdot 10^3$ $\downarrow$}
				&{\it Initial}	&12.290					&7.770					&17.395\\
				&\skmr  				&10.761					&\textbf{6.517}	&16.091\\
				&\skmc 					&\textbf{9.524}	&6.737					&\textbf{12.585}\\
				\cline{2-5}
					\Xhline{2\arrayrulewidth}
				
				\multirow{3}{*}{Normal Consistency $\uparrow$}
				&{\it Initial}	&76.76					&80.49					&63.11\\
				&\skmr 					&80.39					&\textbf{84.43}	&68.67\\
				&\skmc 					&\textbf{81.00}	&83.10					&\textbf{70.49}\\
				\cline{2-5}
					\Xhline{2\arrayrulewidth}
		\end{tabular}}
        \end{tabular}\\
        Cars \hspace{4cm}  \hspace{4cm} Chairs
	\end{center}
\caption{\label{tab:refinement_suppl} \small \textbf{Cars and Chairs.}
	Reconstruction metrics when using the encoding/decoding network trained on \texttt{SketchFD} synthetic sketches of cars and of chairs, and tested on all 3 datasets. We show \textit{initial} results before refinement and then using our two refinement methods. Note that \skmc{} does better than \skmr{}  on the styles it has {\it not} been trained for, indicating a greater robustness to style changes.
}
\end{table*}

In the main paper, in Sec.~\ref{subsec:choosing_best_method} we present a comparison of \skmr{} and \skmc{} approaches when applied on networks trained on \texttt{Suggestive} synthetic sketches, and tested on all 3 datasets. In Tab.~\ref{tab:refinement_suppl} we present the same comparison, but this time for encoder/decoder pairs trained on \texttt{SketchFD}. Again, \skmc{} appears to be more robust to style change and performs better than \skmr{} on datasets the latter has not been trained on.

\subsubsection{Gradients and Sensitivity to Thin Components}

In Fig.~\ref{fig:refinement}, we demonstrate how \skmc{} is more sensitive to thin shape components such as chair legs. Indeed, removing a thin component only affects $\mathcal{L}_{F/B}$ of a few pixels, whereas $\mathcal{L}_{CD}$ takes into account the distance and spatial extent of the deformation.


\begin{figure*}
	\begin{center}
	 \begin{overpic}[clip, trim=0cm 0.0cm 15.6cm 0cm,width= .95\textwidth]{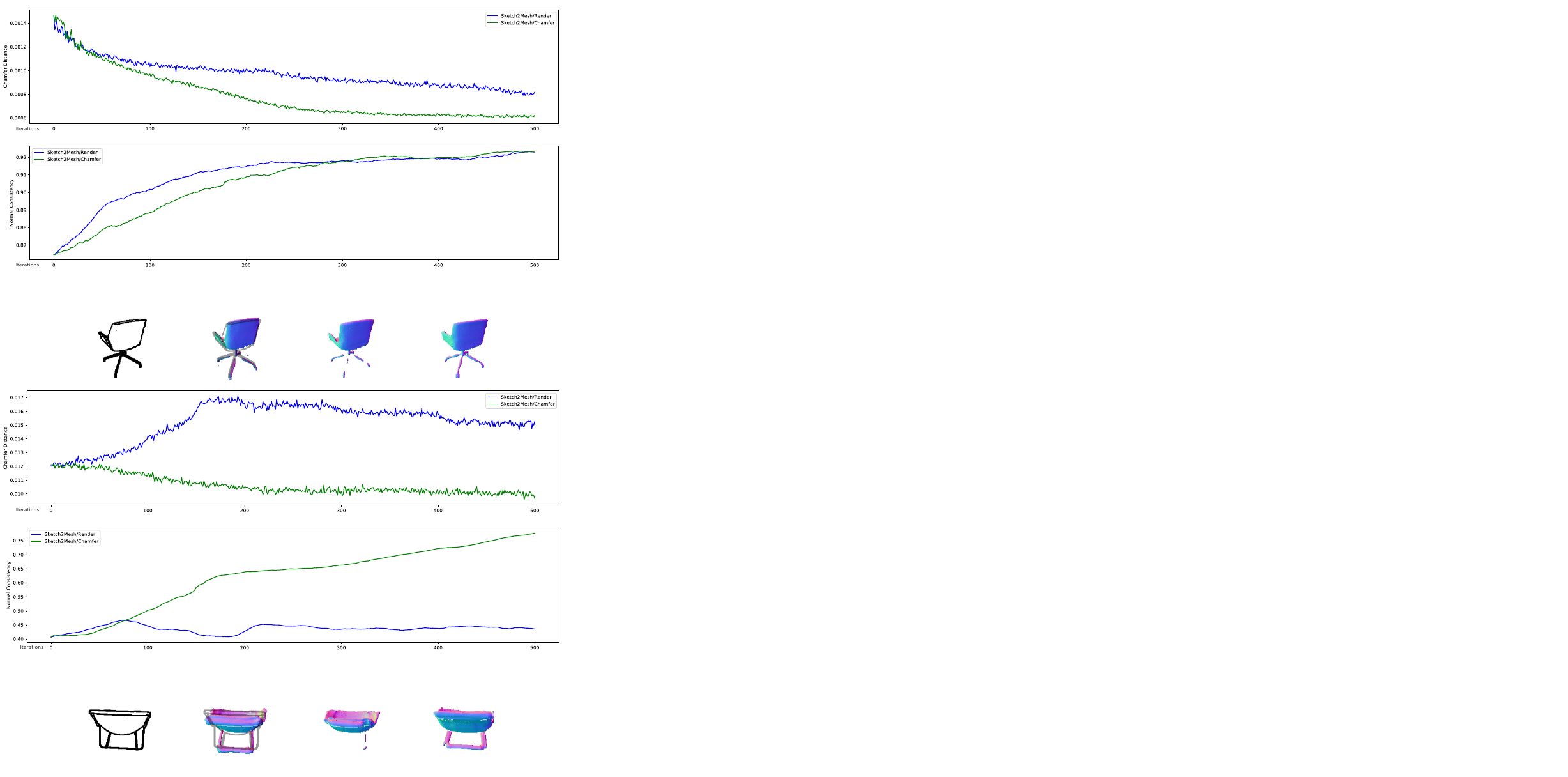}
	 	 	\put(14,12){\small{input}}
	 	 	\put(26,12){\small{reconstruction}}
	 	 	
	 	 	\put(39,12){\small{\skmr{} }}
	 	 	\put(53,12){\small{\skmc{} }}

	 	 	\put(14,62){\small{input}}
	 	 	\put(26,62){\small{reconstruction}}
	 	 	
	 	 	\put(39,62){\small{\skmr{} }}
	 	 	\put(53, 62){\small{\skmc{} }}

 	\end{overpic}
	\end{center}
	\vspace{-3mm}
		 \caption{\textbf{Refinement.}  \skmc{} is more sensitive to thin shape components such as chair legs with respect to \skmr{} : this is due to the nature of the loss, penalizing chamfer distance between silhouettes, rather than per-pixel discrepancies. Best seen digitally, zoomed in. }
	\label{fig:refinement}
\end{figure*}

\comment{
\begin{figure*}
\begin{center}
 	\begin{overpic}[clip, trim=0cm 7.0cm 0cm 0cm,width= .95\textwidth]{figs/prosketch.pdf}

 	\put(5,36){\small{input}}
 	\put(18,36){\small{reconstruction}}
 	
 	\put(33,36){\small{silhouette gap}}
 	\put(49,36){\small{silhouette refinement}}
 	
 	\put(69,36){\small{contour gap}}
 	\put(85,36){\small{contour refinement}}
 	
 \end{overpic}
\end{center}
\vspace{-3mm}
   \caption{\textbf{reconstructions on ProSketch.} \ER{Compare all methods on this, harder because different cameras, different style...}. }
\label{fig:prosketch}
\end{figure*}
}

\subsubsection{Differentiable Rasterization Hyperparameters}

\begin{figure*}
	\begin{center}
	 \begin{overpic}[clip, trim=0cm 0.0cm 15.6cm 0cm,width= .95\textwidth]{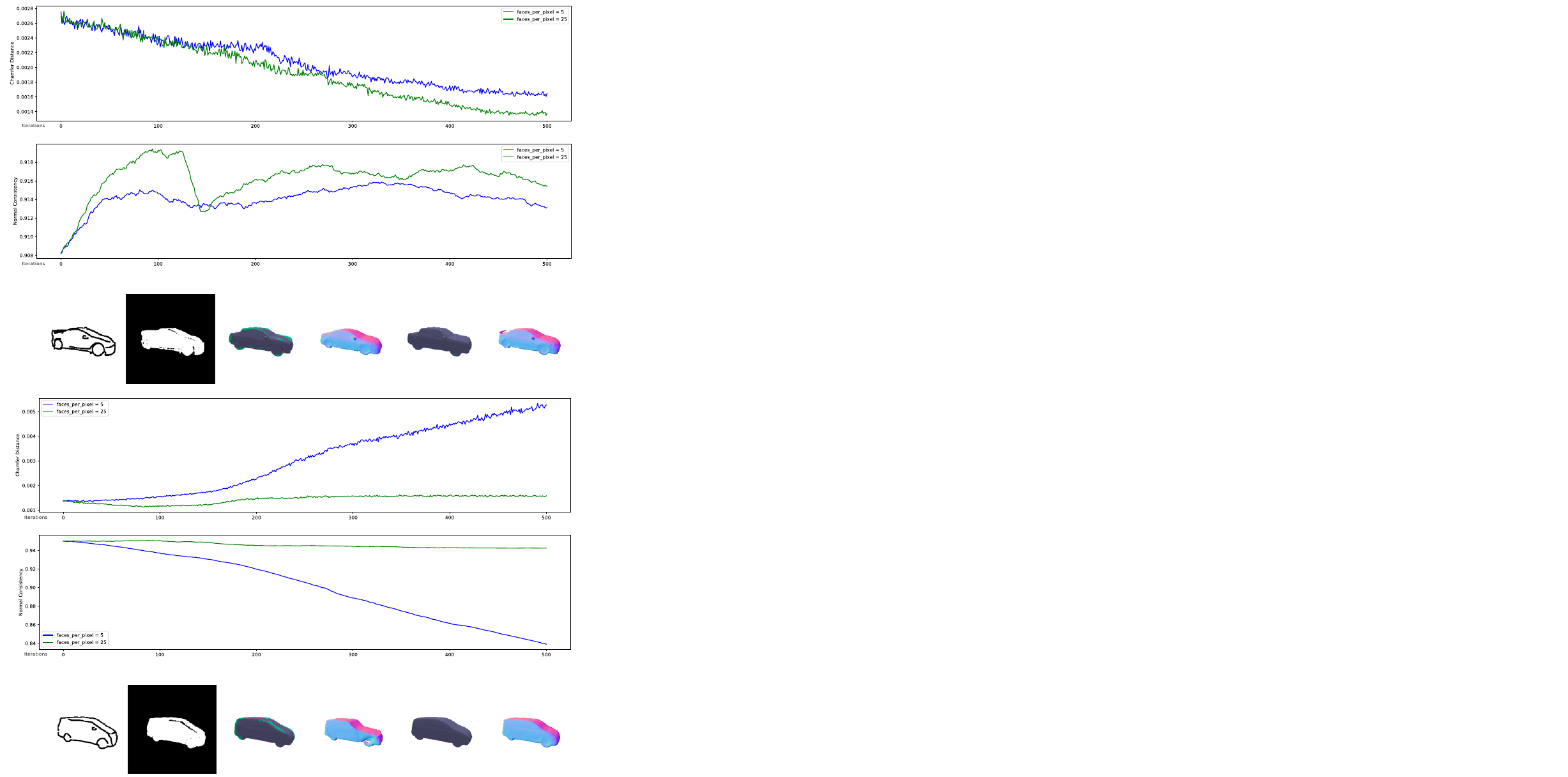}
	 	 	\put(8,13){\small{input}}
	 	 	\put(16,13){\small{predicted silhouette}}
	 	 	
			\put(38,13){\polygon(12,1.5)(12,-10)(-9,-10)(-9,1.5)}
			\put(32.5,13){\small{\texttt{faces\_per\_pixel=5}}}
			\put(29.75,11.2){\small{gradients at}}
	 	 	\put(31,10){\small{iter = 0}}
			\put(42.5,11.2){\small{shape at}}	
	 	 	\put(42,10){\small{iter = 500}}
	 	 	
			\put(61,13){\polygon(12,1.5)(12,-10)(-9,-10)(-9,1.5)}
			\put(54.5,13){\small{\texttt{faces\_per\_pixel=25}}}
			\put(52.75,11.2){\small{gradients at}}
	 	 	\put(54,10){\small{iter = 0}}
			\put(65.5,11.2){\small{shape at}}	
	 	 	\put(65,10){\small{iter = 500}}

	 	 	\put(8,63.5){\small{input}}
	 	 	\put(16,63.5){\small{predicted silhouette}}
	 	 	
			\put(38,63.5){\polygon(12,1.5)(12,-10)(-9,-10)(-9,1.5)}
	 	 	\put(32.5,63.5){\small{\texttt{faces\_per\_pixel=5}}}
			\put(29.75,61.7){\small{gradients at}}
	 	 	\put(31,60.5){\small{iter = 0}}
			\put(42.5,61.7){\small{shape at}}	
	 	 	\put(42,60.5){\small{iter = 500}}
	 	 	
			\put(61,63.5){\polygon(12,1.5)(12,-10)(-9,-10)(-9,1.5)}
			\put(54.5,63.5){\small{\texttt{faces\_per\_pixel=25}}}
			\put(52.75,61.7){\small{gradients at}}
	 	 	\put(54,60.5){\small{iter = 0}}
			\put(65.5,61.7){\small{shape at}}
	 	 	\put(65,60.5){\small{iter = 500}}

 	\end{overpic}
	\end{center}
	\vspace{-5.5mm}
		 \caption{\textbf{Silhouette Alignment.} We compare different settings of pytorch3D \cite{ravi2020accelerating} on two human-drawn car samples. Surface gradients are shown in color (green = intrusion along the surface normal, purple = extrusion).When considering a lower number of triangles for backpropagation of the rasterization process, gradients are more influenced by erroneous silhouette predictions, making refinement less effective (top) or even detrimental (bottom). Best seen digitally, zoomed in. }
	\label{fig:pytorch3D}
\end{figure*}

In Figure \ref{fig:pytorch3D}, we show that the the choice of hyper-parameters in the differentiable rendering process 
can deeply influence the refinement behavior of \skmr{}, particularly when the predicted binary masks are not accurate.
Specifically, we investigate the importance of parameter \texttt{faces\_per\_pixel}, controlling how many triangles are used for backpropagation within each pixel: using a higher number of triangles will result in smoother gradients, as binary mask information is back-propagated to more faces. As depicted in Figure \ref{fig:pytorch3D} this is particularly beneficial when predicted binary masks are not accurate or noisy, and results in more accurate reconstructions. In practice, we set \texttt{faces\_per\_pixel=25} in our experiments.


\comment{
\begin{figure*}
\begin{center}
 	\begin{overpic}[clip, trim=0cm 7.0cm 0cm 0cm,width= .95\textwidth]{figs/prosketch.pdf}

 	\put(5,36){\small{input}}
 	\put(18,36){\small{reconstruction}}
 	
 	\put(33,36){\small{silhouette gap}}
 	\put(49,36){\small{silhouette refinement}}
 	
 	\put(69,36){\small{contour gap}}
 	\put(85,36){\small{contour refinement}}
 	
 \end{overpic}
\end{center}
\vspace{-3mm}
   \caption{\textbf{reconstructions on ProSketch.} \ER{Compare all methods on this, harder because different cameras, different style...}. }
\label{fig:prosketch}
\end{figure*}
}

\end{document}